\documentclass[journal]{IEEEtran}

\ifCLASSINFOpdf
  \usepackage{graphicx}
   \DeclareGraphicsExtensions{.pdf,.jpeg,.png}
\else
\fi

\usepackage{amsmath,amssymb}
\usepackage{algorithmic}

\usepackage{algorithm}

\usepackage[absolute]{textpos}
\newcommand{\copyrightstatement}{
    \begin{textblock}{13.17}(1.36, 15.22)    
         \noindent
         \footnotesize
         \copyright ~2020 IEEE.  Personal use of this material is permitted.  Permission from IEEE must be obtained for all other uses, in any current or future media, including reprinting/republishing this material for advertising or promotional purposes, creating new collective works, for resale or redistribution to servers or lists, or reuse of any copyrighted component of this work in other works.
    \end{textblock}
}

\usepackage{textcomp}
\newtheorem{definition}{Definition}
\usepackage{float}
\usepackage{url}

\usepackage[table,xcdraw, pdftex]{xcolor}
\usepackage{caption}
\usepackage[skip=2pt,font=footnotesize]{caption}
\usepackage{multirow}

\usepackage{enumitem}

\usepackage{csquotes}

\usepackage{longtable}

\usepackage{dblfloatfix}

\begin{document}
\copyrightstatement
%
\title{Adversarial System Variant Approximation to Quantify Process Model Generalization}
%
%
%

\author{Julian~Theis,~\IEEEmembership{Graduate Student Member,~IEEE,}
        and~Houshang~Darabi,~\IEEEmembership{Senior~Member,~IEEE}
\thanks{J. Theis and H. Darabi are with the Department
of Mechanical and Industrial Engineering, University of Illinois at Chicago, 
842 West Taylor Street, Chicago, IL 60607, United States. H. Darabi is the corresponding author. 
E-mail: \{jtheis3, hdarabi\}@uic.edu}
}

\maketitle


\begin{abstract}
In process mining, process models are extracted from event logs using process discovery algorithms and are commonly assessed using multiple quality metrics. While the metrics that measure the relationship of an extracted process model to its event log are well-studied, quantifying the level by which a process model can describe the unobserved behavior of its underlying system falls short in the literature. In this paper, a novel deep learning-based methodology called Adversarial System Variant Approximation (AVATAR) is proposed to overcome this issue. Sequence Generative Adversarial Networks are trained on the variants contained in an event log with the intention to approximate the underlying variant distribution of the system behavior. Unobserved realistic variants are sampled either directly from the Sequence Generative Adversarial Network or by leveraging the Metropolis-Hastings algorithm. The degree by which a process model relates to its underlying unknown system behavior is then quantified based on the realistic observed and estimated unobserved variants using established process model quality metrics. Significant performance improvements in revealing realistic unobserved variants are demonstrated in a controlled experiment on 15 ground truth systems. Additionally, the proposed methodology is experimentally tested and evaluated to quantify the generalization of 60 discovered process models with respect to their systems. 
\end{abstract}


%
\IEEEpeerreviewmaketitle

\section{Introduction}\label{sec:introduction}
Process mining is a comparatively young research discipline that delights itself on ever-increasing popularity with applications in multiple domains such as Healthcare \cite{Mans2015}, Manufacturing \cite{Theis2019c}, Robotic Process Automation \cite{geyer2018process}, Human-Computer Interaction \cite{Theis2019b}, and Simulation \cite{Camargo2019AutomatedLogs}. Commonly, process mining techniques are leveraged to analyze a system. The system is observed during runtime and process steps are recorded sequentially. Recordings are then used to discover process models that are supposed to provide insights into the underlying system, i.e. to describe its behavior. An illustrative example is a complex manufacturing plant of a given product. Such a system usually requires machines to be filled with raw materials, components to be moved and assembled, and final products to be placed on pallets. Examples of corresponding observable process steps are \textit{conveyor start}, \textit{start filling pallet}, \textit{stop assembling}, or \textit{stop machine}. In this case, the sequences of steps are recorded and used to develop a process model that describes the behavior of the manufacturing plant.

In process mining, four metrics are used to assess the quality of discovered process models: \textit{fitness}, \textit{precision}, \textit{simplicity}, and \textit{generalization} \cite{vanderAalst2016conformance}. The first three metrics have received a lot of attention in the literature over the last decade. However, \textit{generalization} is a comparatively new and fairly untouched metric. Dongen et al. \cite{prec_def1} describe generalization as \enquote{how well the model explains unobserved system behavior}. Similarly, Syring et al. \cite{Syring2019} describe generalization as a metric that is \enquote{concerned with quantifying how well a process model generalizes to behavior that is possible in the (...) process but was never observed}. It is obvious that assessing the generalization metric is difficult since unobserved behavior is naturally unknown. 

Process models are used to draw conclusions about the underlying system and future behavior \cite{Rehse2018}. Generalization, as an assessment metric, cannot be ignored in evaluating process models, because a process model that is solely evaluated on \textit{fitness}, \textit{precision}, and \textit{simplicity} metrics might represent the observed recordings only \cite{Syring2019}, and hence overfits. Such a process model is inappropriate to derive conclusions about the system behavior. Recent research studies \cite{Syring2019, Janssenswillen2018}  have shown that state-of-the-art methods fail in assessing the extent to which a process model generalizes unobserved behavior. Hence, the measurement of generalization remains unsolved.

This paper addresses the problem of quantifying generalization by utilizing Generative Adversarial Networks (GANs) \cite{Goodfellow}. GANs approximate data distributions from a given sample dataset. GANs have been extensively applied in the field of computer vision  \cite{Karras2018}, and are capable of learning true underlying distributions of data \cite{Goodfellow, Arora2017}. 

This paper is one of the first in applying GANs in the context of process mining by proposing and statistically testing a novel methodology to overcome the difficulty of assessing the quality metric of generalization. In particular, a Sequence Generative Adversarial Network (SGAN) is proposed to obtain a neural network that approximates the underlying system behavior. This neural network is trained on observed behavior only and is used to reveal the system\textquotesingle s unobserved behavior. The generalization quality can then be assessed by comparing the unobserved behavior modeled by the process model with the unobserved behavior modeled by the neural network. The approach is called \textbf{\underline{A}}dversarial System \textbf{\underline{Va}}rian\textbf{\underline{t}} \textbf{\underline{A}}pp\textbf{\underline{r}}oximation (AVATAR). The effectiveness of the proposed methodology is evaluated and statistically tested in a controlled experiment utilizing $15$ systems. 

The paper is structured as follows. The related work is discussed in Section \ref{sec:related-work}. The fundamental preliminaries are provided in Section \ref{sec:preliminaries} followed by the proposed AVATAR methodology in Section \ref{sec:methodology}. An experimental evaluation using $15$ ground truth systems and comprehensive result interpretations are reported in Section \ref{sec:Experimental-Evaluation}. Finally, Section \ref{sec:conclusion} concludes the paper and provides future research directions. A notations overview and further detailed results can be found in the Appendix.

\section{Related Work}\label{sec:related-work}
The importance of generalization, i.e. measuring how well a process model represents the underlying system, is well-known \cite{vanderAalst2016conformance, Syring2019, generalization}. This stems from the difficulty of deriving the unknown system from a observation sample of limited size. To date, only a few approaches measuring generalization have been proposed.

Van der Aalst et al. \cite{generalizationb} suggested a method which is based on trace alignments. The measure builds on the probability of observing a new path when visiting the same process state the next time. When the average probability is low, the generalization is assumed to be high. However, this metric ignores whether a generated path is realistic, but focuses on the structural properties of the process model. Thus, this approach does not provide insights into the actual underlying system.

Buijs et al. \cite{generalization} developed a measurement that assumes that a process model is likely to represent the underlying system if all parts of the model are frequently used. The method quantifies how often certain parts of a model are used when replaying an event log. Similar to \cite{generalizationb}, this method does not consider if unseen but replayable behavior is likely to be realistic and focuses on structural assumptions only. Therefore, it fails to measure the extent to which a process model represents the behavior of the system.

A further approach has been proposed by vanden Broucke et al. \cite{VandenBroucke2014} utilizing artificial negative events. This method, which is called \textit{Negative Event Generalization}, is based on allowed and disallowed generalizations that can be replayed by a process model. Therefore, generalized events refer to process steps that were not recorded in a log and not considered as negative events in a given context. Such events are presumably realistic system events. However, negative events are generated using an algorithm that checks if certain negative events do occur and were preceded with a similar execution history observed in the log. Hence, this method does not generalize unobserved behavior to full extent by allowing comparatively small variations of the recordings in the event log.

Similarly, van Dongen et al. introduced an approach that is based on anti-alignments \cite{prec_def1}. Anti-alignments are alignments that significantly differ from realistic ones. It is assumed that a likely possible behavior is directly reasoned by measuring the distance to anti-alignments. Therefore, process models that add new behavior without introducing new states are preferred and considered well-generalizing. However, this method ignores whether a behavior has not been observed in an event log \cite{Syring2019}.

To accurately argue how well a process model generalizes, full knowledge about the underlying system must be given. Hence, the existence of a \textit{system model} in a controlled experiment is assumed such that synthetic event logs can be generated \cite{Syring2019}. Measuring generalization can then be reduced to the similarity between discovered process models and the system model \cite{Buijs2014, Janssenswillen2016} using standard notations. Jannsenswillen and Depaire \cite{Janssenswillen2018} concluded that none of the existing methods do reliably measure generalization as required. A similar outcome is reported in \cite{Syring2019}. This shows the importance of this paper that develops a novel generalization approach by approximating the system model from an event log. The proposed AVATAR methodology leverages the properties of GANs which can theoretically unveil underlying true data distribution \cite{Goodfellow}. 

\section{Preliminaries}\label{sec:preliminaries}
\subsection{Logs}\label{sec:preliminaries-logs}
The subsequent definitions are partially based on \cite{vanderAalst2016eventlogs, theis2019dream}. Let $S$ be a system. An event $a \in \mathcal{A}$ is an instantaneous change of the state of $S$ where $\mathcal{A}$ is the finite set of all possible events. The cardinality of $\mathcal{A}$ is denoted by $|\mathcal{A}|$ which corresponds to the size of the set. During the system runtime, an event can occur multiple times. For every occurrence of an event, an event instance $E$ is recorded which is a vector with at least two elements: the label of the corresponding event and the timestamp of occurrence. Consequently, $e_1$ is the first element of the vector $E$ and denotes the label of the recorded event. Further optional elements of an event instance, called metadata, can be other event attributes, such as \textit{resources}, \textit{people}, \textit{costs}, etc. In this work, only the mandatory elements of event instances are utilized. Since events are instantaneous and the point probabilities in continuous probability distributions are zero, two event instance's timestamps cannot be equal. 

A \textit{trace} $c \in \mathcal{C}$ is a finite and chronologically ordered sequence of event instances where $\mathcal{C}$ is the infinite set of all possible traces. Let $\gamma(c)$ be a function that returns the number of event instances of $c$, i.e. its length. The $i$th event instance of $c$ is denoted by $c^i$.

A \textit{variant} $v \in \mathcal{V}$ is a sequence of events where $\mathcal{V}$ refers to the infinite set of variants. Let $\omega: \mathcal{C} \xrightarrow{} \mathcal{V}$ be a function which maps traces to variants such that for a given $c$ its variant satisfies the subsequent condition.
\begin{equation}
\forall_{1 \leq i \leq \gamma(c)}:v^{i} = c^{i}_{1}
\end{equation}

An \textit{event log} $\mathcal{L}$ is a set of traces $\mathcal{L} \subseteq \mathcal{C}$. A \textit{variant log} $\mathcal{L}^*$ is a sample of variants of size $|\mathcal{L}|$ and defined such that $\forall_{c \in \mathcal{L}}: \omega(c) \in \mathcal{L}^*$ and $\forall_{v \in \mathcal{L}^*} \exists_{c \in \mathcal{L}}: v = \omega(c)$. Furthermore, a \textit{unique variant log} $\mathcal{L}^+$ is the set of variants contained in $\mathcal{L}^*$, i.e. $\forall_{v \in \mathcal{L}^*} : v ~\exists!~ \mathcal{L}^+$. The set of all variants of $S$ is denoted by $\mathcal{V}_S$.

Let $R$ be a random variable of $S$ that takes on variants $v \in \mathcal{V}$ and follows a probability density denoted by $\mathbb{P}$. When observing $S$ and recording an event log for an infinite period of time $t \xrightarrow{} \infty$, the relative frequency of each $v \in \mathcal{L}^*$ will follow $\mathbb{P}$.

\subsection{Petri Net}\label{sec:preliminaries:playout}
A Petri net ($PN$) is a mathematical modeling technique. It is commonly used to represent process models. A $PN$ can be visualized as a directed graph in which nodes correspond to places and transitions. Each place can hold tokens which define the state of the $PN$. Tokens move from one place to another by executing transitions. Such transitions correspond to events of a system. This paper refers to \cite{vanderAalst2016petri, PNBasic} for formal $PN$ introductions and definitions. 
A $PN$ is at any given time in a certain state defined by its tokens, called a marking. Variants can be simulated by moving from one marking of the $PN$ to another which is called \textit{playout}. The set of variants that a $PN$ can play out is denoted by $\mathcal{V}_{PN}$. $PN$ process models can be algorithmically extracted from an event log. Such an extraction is called \textit{Process Discovery}. Two relevant state-of-the-art process discovery algorithms are introduced in Section \ref{sec:preliminaries:processdiscovery}.

\subsection{Relations between Event Logs, Process Models, and a System}
\begin{figure}[h]
    \centering
    \includegraphics[width=200pt]{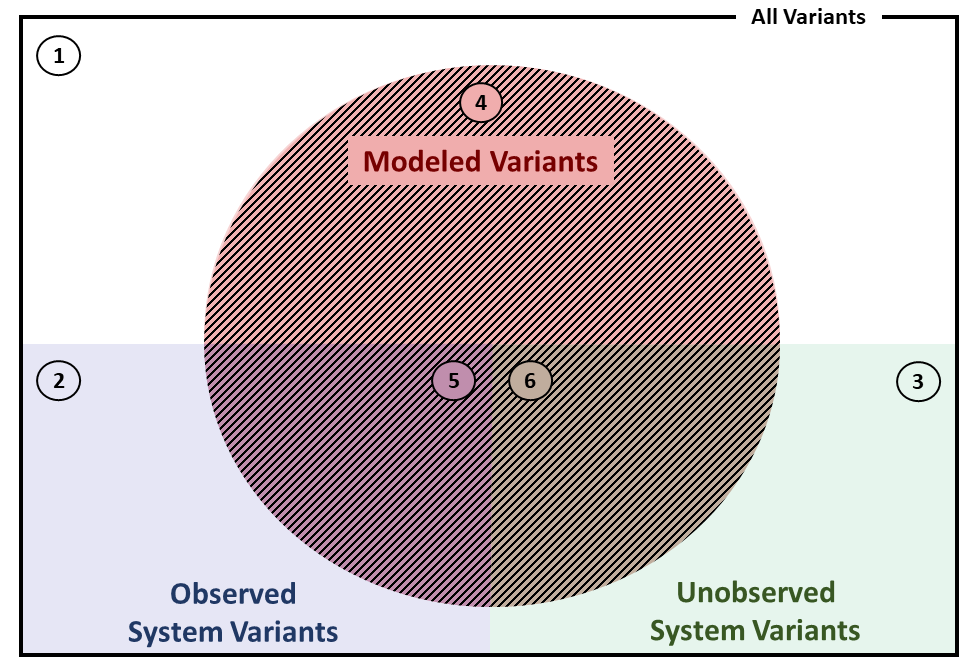}
  \caption{This Venn diagram is derived from \cite{Buijs2014} and illustrates all possible overlaps between realistic and unrealistic, observed and unobserved, and modeled variants in the space of $\mathcal{V}$.}
  \label{fig:venn_diagrams}
\end{figure}

Given a $PN$ discovered from $\mathcal{L}$ that has been recorded over a finite time $t$, its corresponding $\mathcal{L}^+$ can be related to $S$ and $PN$ \cite{generalization, Buijs2014}. $\mathcal{L}^+$ is understood as the observed realistic variants such that $\mathcal{L}^+ \subseteq \mathcal{V}_S$. Consequently, there might exist a subset of realistic variants that have not been observed, denoted by $\mathcal{V}_u$, such that $\mathcal{V}_S = (\mathcal{L}^+ \cup \mathcal{V}_u$) and ($\mathcal{L}^+ \cap \mathcal{V}_u) = \emptyset$. 
A $PN$ extracted from $\mathcal{L}$ models a set of variants denoted by $\mathcal{V}_{PN}$ and is supposed to approximate $\mathcal{V}_S$. Therefore, $\mathcal{V}_{PN}$ is an approximator of the system variant set denoted by $\hat{\mathcal{V}}_{S}$. All variant sets can be related using the Venn diagram shown in Figure \ref{fig:venn_diagrams}. The Venn diagram illustrates the overlaps between all possible variant sets: the observed realistic variants $\mathcal{L}^+$ from $\mathcal{L}$, the realistic unobserved ones, i.e. $\mathcal{V}_{u}$ that correspond to the variants of $S$ that have not been recorded in $\mathcal{L}^*$, and the set of variants that are modeled using a $PN$, i.e. $\mathcal{V}_{PN}$. Each enumerated area can be described as follows:

\begin{enumerate}
  \item \textbf{Unrealistic and unmodeled variants} $(\mathcal{L}^+ \cup \mathcal{V}_{u} \cup \mathcal{V}_{PN})^\complement$\\
  This area of the Venn diagram contains all variants that are neither modeled nor realistic. 
  
  \item \textbf{Observed and unmodeled variants} $(\mathcal{L}^+ \cap \mathcal{V}_{PN}^\complement)$\\
  This area represents all realistic variants that are observed in $\mathcal{L}^*$, but are not contained in $\mathcal{V}_{PN}$.
  
  \item \textbf{Unobserved and unmodeled variants} $(\mathcal{V}_{u} \cap \mathcal{V}_{PN}^\complement)$\\
  This area corresponds to all realistic variants that have not been observed in $\mathcal{L}$ and are not contained in $\mathcal{V}_{PN}$.
  
  \item \textbf{Unrealistic and modeled variants} $(\mathcal{V}_{PN} \cap (\mathcal{L}^{+\complement} \cup \mathcal{V}_{u}^\complement))$\\
  This area represents the variants that $PN$ models but are not realistic.
  
  \item \textbf{Modeled and observed variants} $(\mathcal{L}^+ \cap \mathcal{V}_{PN})$\\
  This area equals to the realistic variants that have been observed and are modeled by $PN$. 
  
  \item \textbf{Modeled and unobserved variants} $(\mathcal{V}_{u} \cap \mathcal{V}_{PN})$\\
  This area corresponds to the unobserved realistic variants that are modeled by $PN$.
\end{enumerate}

\subsection{Model-Log and Model-System Similarity}\label{sec:model-log-system-similarity}
The similarity between a $PN$ and a $\mathcal{L}$ can be measured using two functions \cite{Janssenswillen2018}, as stated in Definition \ref{def:log-fit} and \ref{def:log-prec}.

\begin{definition}\label{def:log-fit}
Log fitness is a function $fit^\mathcal{L} : \mathcal{L} \times PN \xrightarrow{} [0,1]$ that measures how much of the observed traces in $\mathcal{L}$ is modeled by the $PN$. Based on  Figure \ref{fig:venn_diagrams}, this is a quantification of the modeled and observed variants $(\mathcal{L}^+ \cap \mathcal{V}_{PN})$ w.r.t. to $\mathcal{L}^+$.
\end{definition}

\begin{definition}\label{def:log-prec}
Log precision is a function $prec^\mathcal{L} : \mathcal{L} \times PN \xrightarrow{} [0,1]$ that quantifies the unobserved traces that are not contained in $\mathcal{L}$, but modeled by the $PN$. Based on Figure \ref{fig:venn_diagrams}, this refers to a measure that quantifies the modeled and unobserved variants $(\mathcal{L}^{+\complement} \cap \mathcal{V}_{PN})$ w.r.t. to $\mathcal{L}^+$.
\end{definition}

Based on the above definitions, a $PN$ reflects $\mathcal{L}$ perfectly if both, $fit^\mathcal{L}$ and $prec^\mathcal{L}$, equal to $1$. Similarly, $PN$ can be associated to $S$ using two functions \cite{Janssenswillen2018}, as introduced in Definitions \ref{def:sys-fit} and \ref{def:sys-prec}.
\begin{definition}\label{def:sys-fit}
System fitness is a function $fit^S : S \times PN \xrightarrow{} [0,1]$ that measures how much of the realistic variants is modeled by the $PN$. Based on Figure \ref{fig:venn_diagrams}, this is a quantification of the modeled and realistic variants $(\mathcal{V}_S \cap \mathcal{V}_{PN})$ w.r.t. to $\mathcal{V}_S$.
\end{definition}

\begin{definition}\label{def:sys-prec}
System precision is a function $prec^S : S \times PN \xrightarrow{} [0,1]$ that quantifies how much of the unrealistic variants is modeled by the $PN$. Based on Figure \ref{fig:venn_diagrams}, this refers to a measure that quantifies the modeled and unrealistic variants $(\mathcal{V}_{PN} \cap \mathcal{V}_S^\complement)$ w.r.t. to $\mathcal{V}_S$.
\end{definition}

Since the underlying system is commonly unknown, it is challenging to assess $fit^S$ and $prec^S$. Generalization metrics are supposed to measure the extent to which a process model reflects the underlying system by considering the $PN$ and $\mathcal{L}$. 

\begin{definition}\label{def:generalization}
Generalization is a function $gen : \mathcal{L} \times PN \xrightarrow{} [0,1]$ that quantifies modeled and observed variants $(\mathcal{L}^+ \cap \mathcal{V}_{PN})$, modeled and unobserved variants $(\mathcal{V}_{u} \cap \mathcal{V}_{PN})$, and unrealistic and modeled variants $(\mathcal{V}_{PN} \cap \mathcal{V}_S^\complement)$ w.r.t. $(\mathcal{L}^+ \cup \mathcal{V}_{u})$.
\end{definition}

For further illustration, the Venn diagram in Figure \ref{fig:venn_diagrams} visualizes an averagely generalizing $PN$ since the model allows for comparatively many unrealistic variants (area four) while allowing only partially realistic variants (area five and six). Measuring the quality metrics of fitness, precision, and generalization is also referred to as \textit{conformance checking} \cite{vanderAalst2016conformance}.

\subsection{Process Discovery Algorithms}\label{sec:preliminaries:processdiscovery}
In this manuscript, two methods to automatically discover a $PN$ from $\mathcal{L}$ are of interest. Both demonstrate state-of-the-art performance in recent benchmark evaluations \cite{pnbenchmark}.

The first method is \textit{Split Miner} (SM) \cite{splitminer2} which is a process discovery technique that guarantees sound labeled $PN$s \cite{vanderAalst2016petri, soundness}. Its objective is the discovery of process models with high $fit^\mathcal{L}$ and $prec^\mathcal{L}$ scores while keeping the structure as simple as possible. To control the discovery process, SM exposes two hyperparameters with a range of $[0,1]$: a frequency threshold $\varepsilon_1$ that controls the filtering process, and $\varepsilon_2$ that manages the detection of parallelism.

The second process discovery technique is the \textit{Fodina} (FO) algorithm \cite{vandenBroucke2017Fodina:Technique}. It is an extension of the \textit{Heuristics Miner} \cite{Weijters2006} with a comparatively higher robustness against noisy data, the capability of duplicate event detection, and an increased configuration option flexibility. The method exposes multiple hyperparameters, three of which having a high impact on the quality of the discovered model: a dependency threshold $\eta_1$, a length-one-loops threshold $\eta_2$, and a length-two-loops threshold $\eta_3$. Each one is defined in the range $[0,1]$. 

\subsection{Generative Adversarial Networks}
A GAN consists of two neural networks, a generator $G$ and a discriminator $D$, that form a minimax game to optimize each other \cite{Goodfellow}. $G$ is trained to generate realistic samples of a given datatype by modeling a target distribution $\mathbb{Q}$ from samples of a random noise distribution $\mathbb{Z}$. $D$ is trained to differentiate between generated and given realistic samples while directing $\mathbb{Q}$ towards the underlying true data distribution $\mathbb{P}$. The standard loss functions to optimize $G$ and $D$ are defined in Equations \ref{eq:loss-G} and \ref{eq:loss-Dp} where $\mathbf{E}$ corresponds to the expected value. 
\begin{equation}\label{eq:loss-G}
\begin{split}
\displaystyle L_{G} = \mathbf{E}_{z\sim \mathbb{Z}} \big[1-D\big(G(z)\big)\big].
\end{split}
\end{equation}
\begin{equation}\label{eq:loss-Dp}
\begin{split}
\displaystyle L_{D} = \mathbf{E}_{x_r\sim \mathbb{P}}\big[1-D(x_r)\big] + \mathbf{E}_{z\sim \mathbb{Z}} \big[D\big(G(z)\big)\big].
\end{split}
\end{equation}

\subsubsection{Sequence GANs}
A special type of GAN is the Sequence GAN (SGAN) in which the  architecture of $G$ is designed to generate samples of discrete sequences. Its main applications are found within the domain of natural language generation. This paper focuses specifically on an SGAN architecture entitled \textit{Relational GANs for Text Generation} \cite{Gan2019} (RelGAN) due to its superior performance in natural language generation compared to the body of existing state-of-the-art SGAN architectures described in \cite{Zhu2018}.

\subsubsection{RelGAN}\label{sec:relgan}
RelGAN combines three distinct characteristics compared to existing SGAN architectures. First, it does not leverage an LSTM-based generator contrary to its sequential intuition due to apparent drawbacks \cite{Gan2019, Fedus2018, Lu2018NeuralBeyond, Bahdanau2015NeuralTranslate}. Second, RelGAN implements a relational memory-based generator $G$ \cite{Santoro2018RelationalNetworks} for better sample quality. Third, it employs relativistic loss functions for improved training of $G$ and $D$.

The core concept of a relational memory is to \enquote{compartmentalize information and learn to compute interactions between} them \cite{Santoro2018RelationalNetworks}. This is achieved with a memory matrix $M$ where each row corresponds to a memory slot. Interactions between those slots are computed using the \textit{self-attention} mechanism \cite{Vaswani2017AttentionNeed}. Each row of the memory $M_t$ at time $t$ contains a compartmentalized piece of information. With $H$ heads, there are $H$ sets of queries, keys, and values that are calculated through three linear transformations, respectively. 

For every head, one obtains a query $Q_t^{h} = M_tW_q^{h}$, keys $K_t^{h} = M_tW_k^{h}$, and values $Y_t^{h} = M_tW_y^{h}$ where $W_q$, $W_k$, and $W_y$ are row-wise shared weight matrices for the queries, keys, and values, respectively. Intuitively, a query refers to the information one is looking for, the key indicates the relevance to the query, and the value is the actual content of the input. The updated memory $M_{t+1}$ is calculated by
\begin{equation}
    \begin{split}
        \displaystyle 
        M_{t+1} = & \big[M_{t+1}^{1}: ... : M_{t+1}^{H}\big] \\
        M_{t+1}^{h} = & \sigma \Big(\frac{M_tW_q^{h}([M_t;x_t]W_k^{h})^T }{\sqrt{d_k}}\Big) \big[M_t;x_t\big]W_y^{h} 
    \end{split}
\end{equation} where $\sigma(\cdot)$ denotes the softmax function which is performed on each row, $d_k$ is the column dimension of the key $K_t^{h}$, $[:]$ denotes the column-wise and $[;]$ the row-wise concatenation, respectively, and $x_t$ corresponds to a new observation at time $t$. As reported in \cite{Gan2019}, the next memory $M_{t+1}$ and output logits $o_t$ of the generator $G$ at time $t$ equal to
\begin{equation}\label{eq:relgan-nextoutput}
    \begin{split}
        \displaystyle 
        M_{t+1} = f_{\theta_1}\big(M_{t+1}, M_t\big), ~ o_t = f_{\theta_2}\big(M_{t+1}, M_t\big)
    \end{split}
\end{equation}
where $f_{\theta_1}$ and $f_{\theta_2}$ are combinations of skip connections, multi-layer perceptron, gated operations and/or pre-softmax linear transformations.

The discriminator applies the Gumbel-Softmax relaxation technique \cite{Gan2019} to overcome the critical issue of training GANs with discrete data and exposes a tunable parameter called \textit{inverse temperature} denoted by $\beta$ that controls the tradeoff between sample diversity and sample quality. Furthermore, the discriminator learns multiple embedded representations for each sample that is independently passed through a convolutional neural network classifier to obtain a classification score. The output of the discriminator is the mean of each embedded representation score, estimating the probability that given training samples are more realistic than generated ones. The approach stems from the argument that $D(x_r)$ should decrease as  $D\big(G(z)\big)$  increases \cite{Jolicoeur-Martineau2018}. $D_r$ denotes a discriminator that is trained with a relativistic objective. The corresponding loss function is defined in Equation \ref{eq:loss-Dr}.
\begin{equation}\label{eq:loss-Dr}
\begin{split}
\displaystyle L_{D_r} = -\mathbf{E}_{(x_r,x_g)\sim (\mathbb{P},\mathbb{Q})}\big[\log\big(\phi(D_r(x_r) - D_r(x_g)\big)\big]
\end{split}
\end{equation}

The generator $G$ is trained similarly by minimizing the loss function described by Equation \ref{eq:loss-Gr} where $\phi$ refers to the sigmoid function \cite{Jolicoeur-Martineau2018}. 
\begin{equation}\label{eq:loss-Gr}
    \begin{split}
    \displaystyle 
        L_{G} = -\mathbf{E}_{(x_r,x_g)\sim (\mathbb{P},\mathbb{Q})}\big[\log\big(\phi(D_r(x_g) - D_r(x_r)\big)\big]
    \end{split}
\end{equation}

\subsection{Metropolis-Hastings Algorithm}\label{sec:metropolis-hastings}
The Metropolis-Hastings (MH) algorithm \cite{metropolis1953equation, hastings1970monte, Chib1995} can generate sequences of random samples which follow a desired distribution. It is a useful technique to sample from a supposedly complex distribution, i.e. a target distribution such as $\mathbb{P}$. The underlying concept of the MH algorithm relies on a Markov chain \cite{ross2014introduction} with a stationary distribution equaling to $\mathbb{P}$. Thus, the objective is to construct a Markov chain with transition probabilities $p(x,y)$, where $p(x,y)$ is the probability of moving from state $x$ to $y$, such that the Markov chain equals to $\mathbb{P}$ in the long run. 
The algorithm requires a transition kernel $\mathbb{N}$, which is a continuous distribution, to move randomly in space to $y$ given a current position $x$, i.e. $\mathbb{N}(y|x)$ is a density on $y$. Consequently, $\int_{x} \mathbb{N}(y|x)dy = 1$. $\mathbb{N}$ is called the proposal distribution.
The underlying Markov chain is initialized as $X_1 = x_1$. For any $t = 1, 2, 3, ...$, one samples $y$ from $\mathbb{N}(y|x_t)$ where $y$ is understood as a proposed value for $x_{t+1}$. The acceptance probability is calculated by
\begin{equation}
\alpha(x_t,y) = \min\bigg(1, \frac{\mathbb{P}(y)\mathbb{N}(x_t|y)}{\mathbb{P}(x_t)\mathbb{N}(y|x_t)}\bigg)
\end{equation}
with which the proposed sample is accepted, i.e. $x_{t+1} = y$. If a sample is rejected, then $x_{t+1} = x_t$.

As elaborated in \cite{Turner2018}, $\alpha(x_t, y)$ can be computed with the ratio of probability densities $p_{D_p}/p_G$ which is available from the output of the standard discriminator $D_p$ of a GAN. By generating one sample per Markov chain, the MH algorithm can be leveraged to produce iid samples, i.e. $\mathbb{N}(y|x) = \mathbb{N}(y)$. Consequently, $\alpha(x_t, y)$ can be calculated as 
\begin{equation}
\alpha(x_t,y) = \min\bigg(1, \frac{D_p(x_t)^{-1} - 1}{D_p(y)^{-1} - 1}\bigg)
\end{equation}
where $D_p(x) = \frac{p_{D_p}(x)}{p_{D_p}(x) + p_{G}(x)}$. A hyperparameter $\kappa$ defines the length of the Markov chain. For a perfect discriminator and as $\kappa \xrightarrow{} \infty$, this recovers the distribution $\mathbb{P}$. 

\section{Methodology}\label{sec:methodology}
\begin{figure}[h]
    \centering
    \includegraphics[width=220pt]{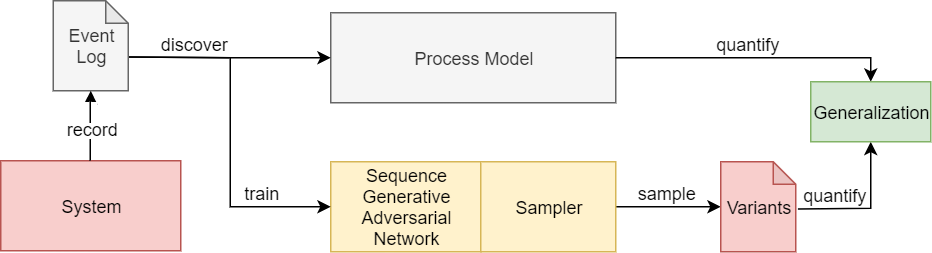}
  \caption{Overview of the AVATAR methodology.}
  \label{fig:methodology-overview}
\end{figure}

This section proposes a methodology called AVATAR to measure the generalization of a $PN$ by approximating $\mathcal{V}_S$ using SGANs. Figure \ref{fig:methodology-overview} provides an overview.

\subsection{Objective}\label{sec:problem-statement}
Based on the definitions in Section \ref{sec:model-log-system-similarity} and the motivation, it is of interest to measure the generalization of a $PN$ discovered by $\mathcal{L}$. However, quantifying the generalization is non-trivial since $\mathcal{V}_{u}$ is naturally unknown. The only information about $S$ is a sample from $\mathbb{P}$ in the form of $\mathcal{L}$ over a finite time $t$. Recent methods aim on quantifying generalization by investigating structural properties of a $PN$ and/or making (probabilistic) assumptions about potential unobserved realistic variants using $\mathcal{L}$. However, this paper follows a different and novel approach leveraging the concept of GANs to address the existing issues. The objective is to approximate $\mathcal{V}_{S}$, and therefore $\mathcal{V}_{u}$, by estimating $\mathbb{P}$ from $\mathcal{L}$ using SGANs. In this way, $\hat{\mathcal{V}}_{S}$ can be utilized to ultimately quantify the generalization of a $PN$ that was discovered from $\mathcal{L}$.

\subsection{Relating SGANs to Process Models}\label{sec:RelatingSGANstoPM}
When discovering a $PN$ from $\mathcal{L}$, state-of-the-art process discovery algorithms consider solely the temporally ordered sequence of event labels, i.e. $\mathcal{L}^*$ of $\mathcal{L}$. Ideally, the discovered $PN$ is capable of playing out variants such that $\mathcal{V}_{PN} = \mathcal{V}_{S}$. If the two sets are equal, the $PN$ is considered perfectly generalizing. Due to the playout property, $PN$s are considered generative models since they can generate variants beyond $\mathcal{L}^*$. Conceptually, this is relatable to a generator of an SGAN which is trained to generate sequences of discrete elements beyond the training set. As such, a $PN$ is theoretically replaceable with a neural network $G: \mathcal{V}_{PN} \approx \mathcal{V}_{G}$ where $\mathcal{V}_{G}$ is the set of variants that $G$ can simulate. In this paper, the objective of $G$ is to generate variants such that $\mathcal{V}_{G} \approx \mathcal{V}_{S}$, since the intention of a discovered $PN$ is to model $S$. $\mathcal{V}_{G}$ and $\mathcal{V}_{PN}$ are both estimators of the variant set of $S$, i.e. $\hat{\mathcal{V}}_S$.

GANs and SGANs have shown impressive results in generalizing underlying data distributions. AVATAR uses these advances to find a neural network model, specifically a parametrized SGAN, that estimates $\mathcal{V}_{S}$ to measure generalization. In other words, an SGAN discovers ideally the real variants of $S$ when trained on a sample. Assuming $\mathcal{V}_{G} \approx \mathcal{V}_{S}$, the generalization of a $PN$ is quantifiable by measuring the similarity between $\mathcal{V}_{PN}$ and $\mathcal{V}_{G}$.

Given $S$ where its $R$ follows $\mathbb{P}$ and given a sufficiently large sample of variants, the SGAN\textquotesingle s generator and discriminator are optimized such that the probability density of $R$ is estimated as $\mathbb{Q}$ where $\mathbb{Q} \approx \mathbb{P}$. In the context of process mining, the required sample to train the SGAN equals $\mathcal{L^*}$ which is a set of realistic variants that were observed over finite time. As explained in the following paragraphs, the methodology requires an SGAN with a generator $G$ and two discriminators: a relativistic and a standard one, denoted by $D_r$ and $D_p$, respectively. Specifically, the standard RelGAN architecture is utilized and extended with a second discriminator. Hereinafter, each SGAN component is related to process mining and the suitability for estimating the unobserved system variants is elaborated to ultimately measure the generalization of a $PN$. 

\subsubsection{Generating Variants}
$G$ is a neural network that generates sequences of events, i.e. variants denoted by $v$, from random noise inputs $z \sim \mathbb{Z}$. 
The function in Equation \ref{eq:maxlen} returns the maximum length of a trace contained in $\mathcal{L}$.
\begin{equation}\label{eq:maxlen}
\displaystyle \mu(\mathcal{L}) = \max\limits_{c \in \mathcal{L}} \gamma(c)
\end{equation}

It is assumed that $\mu(\mathcal{L})$ is the maximum possible variant length of $S$, hence the length of $v \in \mathcal{V}_G$ cannot exceed $\mu(\mathcal{L})$.

The intuition of the relational memory-based $G$ is similar to a system that can be partitioned into smaller subprocesses. These subprocesses correspond to the "compartmentalized information" \cite{Santoro2018RelationalNetworks} of a relational memory. The dependencies which form the system can be referred to as the "interactions between compartmentalized information" \cite{Santoro2018RelationalNetworks}. Practically, this is achieved with $M$ that is used to calculate interactions and updates. A relational memory increases the expressive power of $G$ to generate realistic variants by implicitly identifying the subprocesses and relating their states to $S$. Therefore, a relational memory is suitable for modeling the potential high-modality of $\mathbb{P}$. These modes might range from simple logical rules such as $a_2$ always follows $a_1$ where $\{a_1, a_2\} \subseteq \mathcal{A}$, to complex long-run event patterns. Especially the complex long-run patterns are likely to occur with low probabilities and are consequently rarely or never recorded in $\mathcal{L}$.

The state of each subprocess corresponds to a row of the memory $M_t$ at time $t$. These states and their interdependencies are learned using gradient-based optimization when training $G$. Equation \ref{eq:relgan-nextoutput} illustrates that at time $t$, the next memory's rows and outputs, i.e. events, are combinations of multiple operations denoted by $f_{\theta_1}$ and $f_{\theta_2}$. This can be interpreted as a complex representation of a $PN$ marking and transition firings. Most of the $f_{\theta_1}$ and $f_{\theta_2}$ operations are relatable to $PN$s. Skip connections can express the edges of a $PN$ and gated operations can be inferred as split and join gateways. Firing a certain transition in a $PN$ is equivalent to moving to a new state of $S$ and consequently to new states of the corresponding subprocesses. Each updated row of $M$ represents those state transitions. Thus, $G$ learns implicitly the state transition rules of $S$ which is similar to process discovery.

\subsubsection{Discriminating Variants}
The neural network architecture of $D_r$ led to major performance improvements in natural language generation tasks \cite{Gan2019}. It is therefore uncontested to utilize $D_r$ for discrimination between 
$v \in \mathcal{L}^*$ and $v \in \mathcal{V}_G$.
Its main properties are directly relatable to process mining.

The hypothesis of learning multiple embeddings in $D_r$ is that each embedded representation may capture specific aspects of an input sequence. From a process mining perspective, each embedding implicitly explicates different features that provide insight into subprocess logic or subprocess state interactions of the variant under consideration. Intuitively, a discriminator like $D_r$ that analyzes an input variant from multiple perspectives provides a more comprehensive guiding signal to train $G$ compared to single embedding architectures \cite{Gan2019, Durugkar2019GenerativeNetworks}.

Relativistic loss functions lead to significant performance improvements across different GAN applications, including SGANs. Hence, a relativistic discriminator and generator is preferred to increase the quality of $G$. However, relativistic discriminators are inapplicable to increase the sampling quality using the MH algorithm due to the unavailability of the probability density of an input to be realistic or fake, i.e. $p_{D_p}/p_G$. To leverage both properties, the RelGAN architecture is extended with a second discriminator, denoted by $D_p$. The architectures of both discriminator neural networks are identical, but they are trained with different loss functions. $D_r$ optimizes the relativistic loss described in Equation \ref{eq:loss-Dr} whereas $D_p$ is based on the standard loss by maximizing the probability of correctly classifying observed and generated variants, as described in Equation \ref{eq:loss-Dp}. By extending a RelGAN with a second discriminator, the MH algorithm is applicable to increase the sampling quality, as described in Section \ref{sec:IncreasingSampleQuality}. 

\subsection{Postprocessing}\label{sec:IncreasingSampleQuality}
Sometimes, GANs appear successfully optimized, but $\mathbb{Q}$ is still far from $\mathbb{P}$ \cite{Arora2017}. Turner et al. \cite{Turner2018} developed an approach to overcome this issue on image GANs, as described in Section \ref{sec:metropolis-hastings}. The AVATAR approach transfers the idea from image GANs to process mining SGANs.

By applying the MH algorithm on top of a trained SGAN that generates variants, the likelihood of sampling realistic variants from $\mathbb{P}$ can be increased, as experiments show in Section \ref{sec:Experimental-Evaluation}. Theoretically, a threshold-based filtering of variants generated by $PN$ is realizable without an SGAN, too. However, a threshold value and the likelihood of realistic variants is then only determinable on $\mathcal{L}$ and the structure of $PN$. This goes back to existing generalization measures, as introduced in Section \ref{sec:related-work}. In comparison, when leveraging the MH algorithm on an SGAN, generated variants are evaluated using inferred knowledge beyond $\mathcal{L}$. This stems from the hypothesis that $G$ and $D$ learn characteristics of $\mathbb{P}$. Thus, it is assumable that the MH algorithm can improve the performance of sampling realistic variants contained in $\mathcal{V}_S$.

\subsection{Sampling the Unobserved}\label{sec:Methodology-Sampling}
Discovering $\mathcal{V}_u$ is achieved following two sampling methodologies: naively sampling from $G$ and by utilizing the MH algorithm, denoted by $GAN_{\beta}$ and $MHGAN_{\beta}$, respectively.

Given a trained $G$ that approximates $\mathbb{P}$, variants can be sampled naively from its estimated distribution $\mathbb{Q}$. When sampling $k$ times, $n$ unique variants are observed such that $n \leq k$. The chart of the ordered relative frequencies for each $v$ follows an exponential distribution. When $k$ is a large value, the sorted relative frequency chart of the $n$ unique variants approximates this property. Intuitively, the variants with a high relative frequency are of interest since those are the ones that are modeled confidently by $G$. Consistently, variants with a low relative frequency are modeled with less confidence and are therefore likely to be unrealistic. Thus, the value of $k$ must be reasonably determined during the sampling stage. The union of the sampled $n$ unique variants and $\mathcal{L}^+$ form  $\hat{\mathcal{V}}_S$. Consequently, $\hat{\mathcal{V}}_u = \big(\mathcal{L}^{+^\complement} \cap \hat{\mathcal{V}}_S\big)$ defines the approximated set of unobserved realistic variants. The pseudocode for this sampling procedure is illustrated by Algorithm \ref{algo:naive}.

\begin{algorithm}[h!]
\caption{$GAN_{\beta}$ pseudocode\label{algo:naive}}
\begin{algorithmic}[1]
 \renewcommand{\algorithmicrequire}{\textbf{Input: $G$, $\mathcal{L}^+$, $k$}}
 \renewcommand{\algorithmicensure}{\textbf{Output: $\hat{\mathcal{V}}_S$, $\hat{\mathcal{V}}_u$}}
 \REQUIRE 
 \ENSURE ~

 \STATE $\hat{\mathcal{V}}_S$ = \{ \}
  \FOR {$k = k - 1$ to $0$}
  \STATE $v$ = $G$.generate\_random\_sample( )
  \STATE $\hat{\mathcal{V}}_S$.add($v$)
  \ENDFOR
 \STATE $\hat{\mathcal{V}}_u$ = $\hat{\mathcal{V}}_S$ $-$ intersection($\hat{\mathcal{V}}_S$, $\mathcal{L}^+$)
 \RETURN $\hat{\mathcal{V}}_S$, $\hat{\mathcal{V}}_u$
 \end{algorithmic} 
\end{algorithm}

Rather than naively sampling from $G$ and relying on the described intuition, the MH algorithm can be utilized to estimate $\hat{\mathcal{V}}_S$. A Markov chain of length $\kappa = 500$ is proposed that is initialized with a withheld variant subset $\mathcal{L}^+_e \subseteq \mathcal{L}^+$ to avoid the phenomenon of burn-in \cite{Turner2018}. 
The sampling process terminates as soon as the number of unique generated variants converges. This is controlled using a \textit{patience} hyperparameter $\pi$. More specifically, sampling terminates as soon as none of the $\pi$ most recent generated variants are novel. The pseudocode for this sampling procedure is illustrated by Algorithm \ref{algo:mh}.

\begin{algorithm}[h!]
\caption{$MHGAN_{\beta}$ pseudocode\label{algo:mh}}
\begin{algorithmic}[1]
 \renewcommand{\algorithmicrequire}{\textbf{Input: $G$, $D_p$, $\mathcal{L}^+$, $\mathcal{L}^+_e$, $\pi$, $\kappa$}}
 \renewcommand{\algorithmicensure}{\textbf{Output: $\hat{\mathcal{V}}_S$, $\hat{\mathcal{V}}_u$}}
 \REQUIRE 
 \ENSURE ~
 \STATE $\hat{\mathcal{V}}_S$ = \{ \} \AND $cnt = 0$
 
 \WHILE{$cnt < \pi$}
     \STATE $v_{ref}$ = $\mathcal{L}^+_e$.get\_random\_instance( ) \AND $v_x$ = $v_{ref}$
     \STATE $v_y$ = $G$.generate\_random\_sample( )
     \FOR {$\kappa = \kappa - 1$ to $0$}
        \STATE $\alpha$ = $min(1, (1/D_p(v_x) - 1)/(1/D_p(v_y) - 1))$
        \IF {($\alpha$ > random\_from\_uniform(0,1)}
            \STATE $v_x$ = $v_y$
        \ENDIF
        \STATE $v_y$ = $G$.generate\_random\_sample( )
     \ENDFOR
     \IF {($v_y \neq v_{ref}$ \AND $v_y {\not\in} \hat{\mathcal{V}}_S$)}
        \STATE $\hat{\mathcal{V}}_S$.add($v_y$) \AND $cnt = 0$
     \ELSE
        \STATE $cnt = cnt + 1$
     \ENDIF
\ENDWHILE

 \STATE $\hat{\mathcal{V}}_u$ = $\hat{\mathcal{V}}_S$ $-$ intersection($\hat{\mathcal{V}}_S$, $\mathcal{L}^+$)
 \RETURN $\hat{\mathcal{V}}_S$, $\hat{\mathcal{V}}_u$
 
 \end{algorithmic} 
\end{algorithm}

\subsection{Quantifying Generalization}\label{sec:QuantifyingGeneralization}
Given that $\mathbb{Q} \approx \mathbb{P}$, $\mathcal{V}_u$ is approximable by naively sampling from $G$ or using the MH algorithm with $G$ and $D_p$.
The generalization of a $PN$ is then quantifiable by measuring $(\mathcal{L}^+ \cap \mathcal{V}_{PN})$, $(\hat{\mathcal{V}}_{u} \cap \mathcal{V}_{PN})$, and  $\big(\mathcal{V}_{PN} \cap (\mathcal{L}^{+\complement} \cup \hat{\mathcal{V}}_{u}^\complement)\big)$ w.r.t. $\hat{\mathcal{V}}_S$. The generalization of a $PN$ is suggested to be measured using a synthetically generated event log of the estimated system variants, denoted by $\mathcal{L}_S$ such that 
$|\mathcal{L_S}|$ = $|\hat{\mathcal{V}}_S|$ and $\forall_{c \in \mathcal{L_S}}: \omega(c) \in \hat{\mathcal{V}}_S$. 
The extent to which the $PN$ represents $\hat{\mathcal{V}}_S$ is measured using 
$fit^\mathcal{L}(PN, \mathcal{L}_S^*)$. 
Moreover, $\mathcal{V}_{PN}$ should not contain variants that are not contained in $\hat{\mathcal{V}}_S$. 
This equals to $prec^\mathcal{L}(PN, \mathcal{L}_S^*)$. 
If 
$\hat{\mathcal{V}}_S = \mathcal{V}_S$, 
then 
$fit^S = fit^\mathcal{L}(PN, \mathcal{L}_S^*)$
and
$prec^S = prec^\mathcal{L}(PN, \mathcal{L}_S^*)$. 
Since a generalization score is supposed to be a single value, the harmonic mean of $fit^\mathcal{L}$ and $prec^\mathcal{L}$ on $PN$ and $\mathcal{L}^*_S$ is proposed in Equation \ref{eq:proposed-generalization} as a quantification of generalization. 
\begin{equation}\label{eq:proposed-generalization}
gen(PN,\mathcal{L}_S^*) = 2 * \frac{fit^\mathcal{L}(PN,\mathcal{L}_S^*) * prec^\mathcal{L}(PN,\mathcal{L}_S^*)}{fit^\mathcal{L}(PN,\mathcal{L}_S^*) + prec^\mathcal{L}(PN,\mathcal{L}_S^*)}
\end{equation}

\section{Experimental Evaluation}\label{sec:Experimental-Evaluation}
This section focuses on answering the following four questions to experimentally evaluate the proposed approach.

\underline{Q1}: Does the AVATAR sampling methodology approximate the true number of system variants better than a discovered process model?

\underline{Q2}: Can the AVATAR sampling methodology detect a larger proportion of system variants than a discovered process model?

\underline{Q3}: Does the AVATAR sampling methodology detect a larger proportion of unobserved system variants than a discovered process model?

\underline{Q4}: Does the proposed generalization quantification lead to metric values that are closer to the expected scores? 

The experimental evaluation is based on ground truth systems that are introduced in Section \ref{sec:Experimental-Evaluation:GroundTruth} along with the required metrics in Section \ref{sec:Experimental-Evaluation:Metrics}. Baseline scores for evaluation are established in Section \ref{sec:Experimental-Evaluation:PDandCC} using state-of-the-art process discovery and conformance checking techniques. Finally, Section \ref{sec:Experimental-Evaluation:GAN} investigates the performance of the proposed approach. Section \ref{sec:Experimental-Evaluation:GAN:sampling} provides answers to questions Q1-Q3 that are underscored by statistical tests in Section \ref{sec:Experimental-Evaluation:GAN:stats}. Section \ref{sec:Experimental-Evaluation:Generalization} provides answers to Q4. 

\subsection{Ground Truth Systems}\label{sec:Experimental-Evaluation:GroundTruth}

The ground truth systems on which the evaluation of AVATAR is based, are $15$ artificial $PN$s that have been explained in \cite{Janssenswillen2017ADiscovery} and are publicly available\footnote{ https://github.com/gertjanssenswillen/processquality/}. These models were created based on a system generation algorithm, as described in \cite{Jouck2018}. The method allows different input parameters, such as the distribution for the number of leaf nodes, the distribution for the type of operator nodes, and the probability for silent and duplicate events \cite{Janssenswillen2017ADiscovery} to control the structure of the resulting $PN$. The $15$ ground truth systems are classified as either \textit{moderate complex} or \textit{highly complex} \cite{Janssenswillen2017ADiscovery}, based on the findings in \cite{fitnesseval2}. Process discovery algorithms perform differently when system behavior is complex, such as in real-life applications, compared to more elementary behavior, such as in artificial setups. Therefore, the ground truth systems that are classified as \textit{moderate complex} outline fewer leaf nodes and a smaller proportion of advanced constructs, such as loops and inclusive choices, compared to \textit{high complex} $PN$s.

An overview of the systems and their variants can be found in Table \ref{table:ground-truth-processes}. Playout is performed on each ground truth system to obtain $\mathcal{V}_S$. Loops in each $PN$ structure are controlled such that every place can produce a maximum of three tokens. Hence $\mathcal{V}_S$ is finite. $\mathcal{V}_S$ is randomly split into two mutually exclusive sets $\mathcal{L}^+$ and $\mathcal{V}_u$. $\mathcal{L}^+$ contains $70\%$ of the variants including at least one variant where its size equals the maximum length of the system. $\mathcal{V}_u$ contains the remaining $30\%$ of variants. For every $S$, $\mathcal{L}$ is generated such that $|\mathcal{L}| = |\mathcal{L}^+|$ and $\forall_{c \in \mathcal{L}}: \omega(c) \in \mathcal{L}^+$.

\begin{table}[ht]
\centering
\resizebox{250pt}{!}{%
\begin{tabular}{|c|c|c|c|c|c|c|c|}
\hline
\rowcolor[HTML]{C0C0C0} 
Identifier & Name in \cite{Janssenswillen2017ADiscovery} & Complexity & \textbf{$|\mathcal{A}|$} & \textbf{$\mu(\mathcal{L})$} & \textbf{$|\mathcal{L}^+|$} & \textbf{$|\mathcal{V}_u|$} & \textbf{$|\mathcal{V}_S|$} \\ \hline
System 1           & PA System 1 5         & moderate            &  11                         & 5                             &   124                        & 54                         & 178                            \\ \hline
System 2           & PA System 2 3         & moderate            &  18                         & 17                            & 206                          &     89                     & 295                            \\ \hline
System 3           & PA System 4 3         & moderate            &   15                        & 20                            &  38                         &      16                    & 54                             \\ \hline
System 4           & PA System 5 5         & moderate            &   18                        & 11                            &   941                        &   403                       & 1,344                          \\ \hline
System 5           & PA System 6 3         & moderate            &   18                        & 19                            &   647                        &       277                   & 924                            \\ \hline
System 6           & PA System 7 7         & moderate            & 14                          & 17                            &    1,531                       & 657                         & 2,188                          \\ \hline
System 7           & PA System 8 3         & moderate            &  18                         & 21                            &   84                        &     36                     & 120                            \\ \hline
System 8           & PA System 9 1         & moderate            & 7                          & 8                             &  176                         &   76                       & 252                            \\ \hline
System 9           & PA System 10 2        & moderate            &   14                        & 17                            &   157                        &   67                       & 224                            \\ \hline
System 10          & PA System 11 3        & moderate            &   15                        & 6                             &   164                        &    70                      & 234                            \\ \hline
System 11          & PB System 5 3         & high                &     14                      & 21                            &    290                       &  125                        & 415                            \\ \hline
System 12          & PB System 1 5         & high                &   14                        & 18                            &   476                        &  204                        & 680                            \\ \hline
System 13          & PB System 2 4         & high                &  15                         & 43                            &   355                        &    152                      & 507                            \\ \hline
System 14          & PB System 3 6         & high                &   11                        & 16                            &   546                        & 234                         & 780                            \\ \hline
System 15          & PB System 4 1         & high                &  15                         & 21                            &  481                         &  207                        & 688                            \\ \hline
\end{tabular}%
}
\caption{Ground truth systems and their details.}
\label{table:ground-truth-processes}
\end{table}

\subsection{Metrics}\label{sec:Experimental-Evaluation:Metrics}
Several different metrics are utilized to evaluate the extent to which discovered $PN$s and SGANs model $\mathcal{V}_S$. $tp$ refers to the true positive rate, i.e. the ratio of the number of realistic variants over the number of generated variants where $\hat{\mathcal{V}}_S$ can be either the result of $PN$ playout or sampling via $GAN_{\beta}$ or $MHGAN_{\beta}$. Consequently, the false positive rate is the ratio of the number of unrealistic variants over the number of generated variants, $fp = 1 - tp$. Furthermore, the ratio of the number of generated realistic variants over the number of realistic variants, the ratio of the number of generated realistic variants over the number of observed variants, and the ratio of the number of generated unobserved and realistic variants over the number of unobserved variants are measured and denoted by $tp_{S}$, $tp_{o}$, and $tp_{u}$, respectively. Furthermore, when holding out variants from $\mathcal{L}^+$, such as $\mathcal{L}^+_e$, the ratio of the number of generated realistic variants over the number of holdout variants is calculated as $tp_e$.
\begin{equation}\label{eq:tp-S}
\begin{split}
    \displaystyle tp_{S} = \frac{|\hat{\mathcal{V}}_S \cap \mathcal{V}_S|}{|\mathcal{V}_S|}, ~ tp_{o} = \frac{|\hat{\mathcal{V}}_S \cap \mathcal{L}^+|}{|\mathcal{L}^+|}\\
    tp_{u} = \frac{|\hat{\mathcal{V}}_S \cap \mathcal{V}_u|}{|\mathcal{V}_u|}, ~ tp_{e} = \frac{|\hat{\mathcal{V}}_S \cap \mathcal{L}^+_e|}{|\mathcal{L}^+_e|}
\end{split}
\end{equation}

\subsection{Process Discovery and Conformance Checking}\label{sec:Experimental-Evaluation:PDandCC}
This subsection introduces the experimental setup of discovering process models from the ground truth systems and checking their conformance to establish a baseline.

\subsubsection{Setup}\label{sec:Experimental-Evaluation:PDandCC:Setup}
SM and FO are used to discover $PN$s from $\mathcal{L}$ for each of the $15$ ground truth systems. Hyperparameter optimization is performed to discover high quality process models. For each of the $15$ ground truth system, $121$ $PN$s are discovered using SM with unique $\varepsilon_1$ and $\varepsilon_2$ pairs and parameter interval steps of $0.1$, and $216$ $PN$s using FO with unique $\eta_1$, $\eta_2$, and $\eta_3$ combinations and parameter interval steps of $0.2$. Two process models per process discovery algorithm and ground truth system are selected by performing conformance checking using existing metrics. The process model which resulted in the highest generalization score is selected for each of the process discovery algorithms SM and FO, denoted by $PN_{SM1}$ and $PN_{FO1}$, respectively. Additionally, two further process models $PN_{SM2}$ and $PN_{FO2}$ are selected which score high log fitness and generalization across the discovered process models on each ground truth system by SM and FO, respectively. The intuition is that a $PN$ with a high log fitness and generalization score supposedly models the underlying system. It should be noted that $PN_{SM2}$ or $PN_{FO2}$ can be the same process models as $PN_{SM1}$ or $PN_{FO1}$. Finally, four ideally different hyperparameter-optimized $PN$s per ground truth system are used for further evaluation.

For every ground truth system, four  process models are selected. For each $PN$, token-based log fitness $fit^\mathcal{L}_t$ \cite{rozinat2008conformance} and the \textit{ETConformance}-based log precision $prec^\mathcal{L}_{etc}$ \cite{Carmona2010AConformance}, alignment-based log fitness $fit^\mathcal{L}_a$ and precision $prec^\mathcal{L}_a$ \cite{VanDongen2016Alignment-basedChecking}, and generalization $gen_{ETM}$ \cite{generalization} are obtained. These conformance measures are the de-facto standard when evaluating process models. Two methods are used to evaluate log fitness and log precision since it is not guaranteed that the process discovery algorithms create sound $PN$s, therefore alignment-based metrics are not applicable. However, token-based conformance scores might be ambiguous \cite{Adriansyah2014} and therefore are eventually unrepresentative. Bold values refer to the metrics that have been optimized during hyperparameter determination. The reported conformance checking evaluations are performed using the Python-based open source process mining library PM4Py \cite{Berti2019}.

Playout is performed for each selected $PN$ to obtain $\mathcal{V}_{PN}$ as an estimator of $\mathcal{V}_S$. When generating $\mathcal{V}_{PN}$, the maximum variant length is set to $\mu(\mathcal{L})$ since it is assumed that the log contains a trace which is equal to the maximum variant length of $S$. Based on those sets, the values for $tp$, $fp$, $tp_S$, $tp_{o}$, and $tp_{u}$ are calculated .

\subsubsection{Results}\label{sec:pdandcc:results}
\begin{figure*}[ht]
    \includegraphics[width=\textwidth]{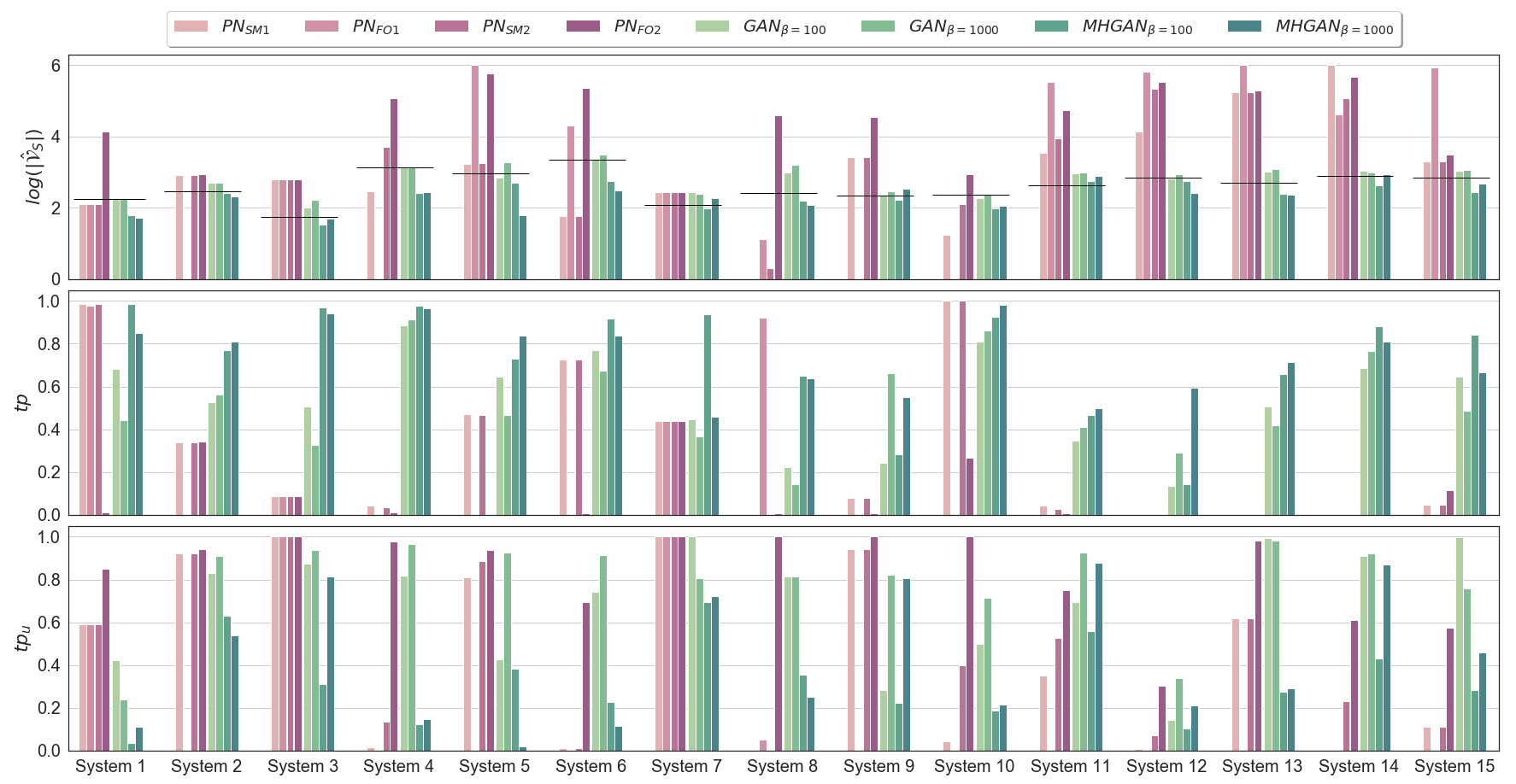}
  \caption{Comparison of the number of $\hat{\mathcal{V}}_S$, $tp$, and $tp_{u}$ of the discovered process models obtained (red) and SGAN sampling models (green) for each of the 15 ground truth systems. Horizontal black lines indicate $|\mathcal{V}_S|$.}
  \label{fig:gan-results}
\end{figure*}  

This subsection reports significant observations. The Appendix contains all detailed results.

$PN_{SM1}$ and $PN_{FO1}$ of \textit{System 1} score comparatively high generalization between $0.828$ and $0.845$. This seems to accurately reflect the generalization of the process models since high $tp$ ratios are obtained. Around $70\%$ and $60\%$ of $\mathcal{L}^+$ and $\mathcal{V}_u$ are covered by $\hat{\mathcal{V}}_S$. However, roughly $50$ variants of $\mathcal{V}_S$ are unmodeled by the $PN$. Hence, the generalization score seems representative, however, the metric was not designed to measure the extent to which unobserved variants are modeled and has been proven to fail, as discussed earlier. $PN_{FO2}$ has a higher fitness, but lower precision and a low generalization score of around $56\%$. This value seems reasonable, too, since $tp$ is comparatively very low. Overall, the model generalizes comparatively well and infers roughly $85\%$ of $\mathcal{V}_u$ .

A generalization misalignment is observed when analyzing the models of \textit{System 2}. $PN_{FO1}$ scores the highest generalization score. However, it cannot playout any realistic variant. Therefore, $PN_{FO1}$ does not generalize at all and the obtained generalization score is misleading. $PN_{FO2}$ performs with higher $tp$, $tp_{S}$, $tp_{o}$, and $tp_{u}$ values than $PN_{SM1}$ or $PN_{SM2}$. However, the SM models score in comparison $0.13$ lower which highlights the misalignment. 

The resulting scores of $PN_{FO1}$ on \textit{System 5} are questionable.
Conformance checking results in a decent token-based fitness, but very low precision. This indicates a spaghetti-like structure of the model which leads to a large $|\hat{\mathcal{V}}_S|$. However, none of the variants in $\hat{\mathcal{V}}_S$ are realistic. Still, this model scores the highest generalization score across the discovered $PN$s, though it is intuitively expected to be close to $0$. 

For \textit{System 11}, four $PN$s are discovered with comparatively high scores of generalization in the range of $0.865$ and $0.915$. However, this is not reflected in the ratios. Out of $415$ possible realistic variants, especially the FO $PN$s tend to model more variants that are not realistic. This results in low $tp$ ratios. 
When optimizing the process discovery for high generalization scores, FO discovers models with low fitness and precision scores which result in unsound $PN$s. Therefore, the generalization score should be rather low than high. However, $PN_{FO2}$ has a much higher fitness and precision, but still a low $tp$ ratio. Though this model plays out more realistic variants, it results in a lower generalization score compared to $PN_{FO1}$. This phenomenon is observed with $PN_{SM1}$ and $PN_{SM2}$, too. $PN_{SM1}$ results in lower conformance checking scores except for generalization compared to $PN_{SM2}$, but generalizes worse according to the true positive ratios. Only $tp$ is improved. 

When performing the experiment on \textit{System 12}, $PN_{FO1}$ and $PN_{SM1}$ are both unsound without fitting variants. However, their generalization scores are above $90\%$ which seems counterintuitive. Nonetheless, $PN_{SM1}$ reproduces certain observed and unobserved variants with a high $fp$ rate. Interestingly, $PN_{FO2}$ results in the same $fp$ rate with much higher $tp_{o}$ and $tp_{u}$ ratios, thus actually generalizing better than $PN_{SM1}$. However, their generalization scores are very different.

Similarly to \textit{System 12}, $PN_{SM1}$ and $PN_{FO1}$ of \textit{System 14} are unsound $PN$s that cannot generate any realistic variant. However, their generalization score is the highest among the discovered $PN$s. According to the true positive ratios, $PN_{SM2}$ and $PN_{FO2}$ generalize with a high $fp$ rate. Their generalization score is above $0.91$, but their true positive ratios vary. $PN_{FO2}$ plays out significantly more variants than $PN_{SM2}$ including more realistic ones.

These results align with the conclusions of \cite{Syring2019, Janssenswillen2018} that the generalization metric implemented in PM4Py does not reflect generalization as desired. Hence, the results emphasize the problem and justify the motivation of this paper.

\subsection{Methodology Evaluation}\label{sec:Experimental-Evaluation:GAN}
\subsubsection{SGAN Training and Sampling}
For each $S$, two extended RelGANs are trained as proposed in Section \ref{sec:methodology}. The first one is trained with $\beta = 100$ to encourage exploration for an improved sample quality \cite{Gan2019}. The second model is trained with the maximum value, $\beta = 1000$, to encourage higher sample diversity. A \enquote{vanilla} relational memory architecture is chosen for $G$ due to observed superior performance over architectural variations \cite{Gan2019}. Furthermore, the head size of the relational memory has been reduced from $512$ to $256$ since the ground truth systems are intuitively less complex than the original application domain of RelGANs. The remaining training parameters are the default values as in \cite{Gan2019}.

$\mathcal{L}^+$ is split into a $90\%$ train set denoted by $\mathcal{L}^+_t$, and a $10\%$ holdout validation set, $\mathcal{L}^+_e$. First, $G$ is pretrained with 100 epochs as described in \cite{Gan2019}. Then, the extended RelGAN is trained for 5,000 adversarial epochs. To obtain $\hat{\mathcal{V}}_S$ and to calculate $tp_e$, 10,000 variants are drawn every 20 epochs from $G$. The best performing model is the one leading to the highest obtained $tp_e$ value and to a comparatively small $|\hat{\mathcal{V}}_S|$. This is a widely adopted approach to train neural networks \cite{earlystopping2}. The determined models are then used to naively sample 10,000 variants ($GAN_{\beta=100}$ and $GAN_{\beta=1000}$) or to sample using the MH algorithm ($MHGAN_{\beta=100}$ and $MHGAN_{\beta=1000}$) to establish four estimators of $\mathcal{V}_S$.

The evaluation procedures are implemented using Texygen \cite{Zhu2018} and Tensorflow \cite{tensorflow}. The process model and event log processing methods are based on PM4Py \cite{Berti2019}. The corresponding source code is publicly available\footnote{https://github.com/ProminentLab/AVATAR}. The training and sampling have been performed on NVIDIA GeForce RTX 2080 GPUs and took 2-3 hours per extended RelGAN and dataset.

\subsubsection{Sampling Results}\label{sec:Experimental-Evaluation:GAN:sampling}
Figure \ref{fig:gan-results} visualizes the number of sampled variants, i.e. $|\hat{\mathcal{V}}_S|$, $tp$, and $tp_u$ for each ground truth system and for each of the four $PN$ and four GAN sampling approaches. The visualization of the number of generated variants show that the SGANs are generating a number of variants closer to the ground truth compared to $PN$s across all datasets. This is even more prominent on the highly complex systems 11-15. Therefore, the answer to Q1 is that the AVATAR sampling methodology indeed approximates the true number of system variants better than process models that are built from state-of-the-art discovery algorithms.

The middle visualization of Figure \ref{fig:gan-results} shows that for most of the systems, the $tp$ values for the GAN sampling approaches are higher than for $PN$s. This is especially prominent for highly complex systems. It can be seen that for $MHGAN_{\beta=100}$ and $MHGAN_{\beta=1000}$ the $tp$ values are increased compared to $GAN_{\beta=100}$ and $GAN_{\beta=1000}$. This indicates that w.r.t. Q2, the AVATAR sampling methodology models a higher proportion of system variants compared to the discovered process models.

The bottom graph visualizes the $tp_u$ ratios and provides information to answer Q3. Most SGANs do not necessarily model more unobserved variants compared to $PN$s. Moreover, when applying the MH algorithm, $tp_u$ values tend to decrease compared to naive GAN sampling. 

A tradeoff between the number of generated variants, $tp$, and $tp_u$ exists and needs to be investigated to answer Q3. When analyzing solely $tp_u$ on systems 7, 8, and 10, it seems that the discovered $PN$s model unobserved variants better than SGANs. However, the corresponding $tp$ is lower than the $tp$ of some SGANs. When considering $tp$ and $tp_u$ combined, SGANs perform significantly better compared to the discovered $PN$s since their $fp$ rate is much lower. This means that the set $\big(\mathcal{V}_{PN} \cap (\mathcal{L}^{+\complement} \cup \mathcal{V}_{u}^\complement)\big)$ is comparatively smaller which is desirable. Moreover, Figure \ref{fig:gan-results} indicates that sampling with MH increases $tp$ and decreases the number of generated variants and $tp_u$ compared to naive GAN sampling. This means that $fp$ is further reduced, however, at the cost of the number of unobserved realistic variants.

\subsubsection{Statistical Significance}\label{sec:Experimental-Evaluation:GAN:stats}
\begin{figure}[!h]
    \centering
    \includegraphics[width=240pt]{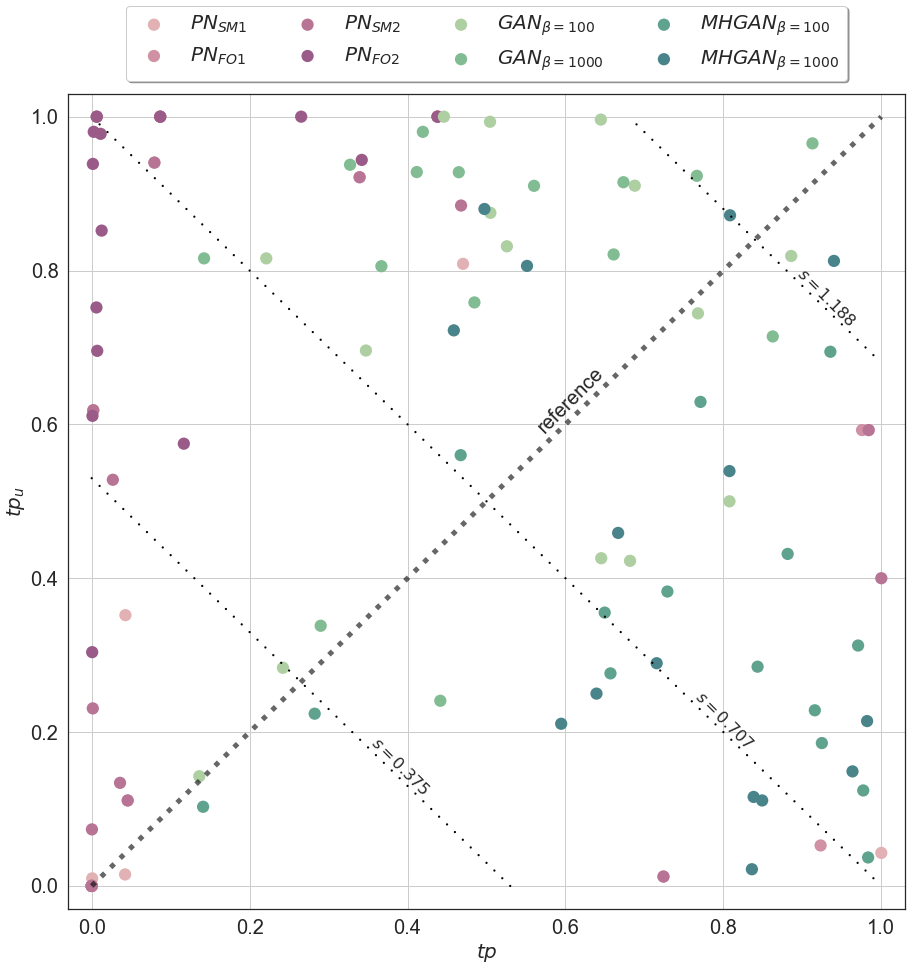}
  \caption{Visualizing $tp_{u}$ over $tp$ for $PNs$s and SGAN sampling models. A score $s$ can be calculated for each model. A higher score indicates a better perceived generalization. The three dashed example lines indicate the scores $s=0.375$, $s=0.707$, and $s=1.188$ in the two-dimensional space.}
  \label{fig:2d_gan_evaluation}
\end{figure}

Ultimately, a generative model is desired such that $\hat{\mathcal{V}}_S = \mathcal{V}_S$. This means that $|\hat{\mathcal{V}}_S|$ should be as close as possible to $|\mathcal{V}_S|$ and $tp$ and $tp_{u}$ are ideally equal to $1.0$. Since most of the values for $|\hat{\mathcal{V}}_S|$ are higher than the corresponding $|\mathcal{V}_S|$, one can combine $tp$ and $tp_{u}$ in a single measure to investigate the tradeoff. Figure \ref{fig:2d_gan_evaluation} illustrates $tp_u$ over $tp$ scores for all $PN$s and SGAN sampling models in a two-dimensional space.

Since highly generalizing models are expected to have a high $tp$ and $tp_{u}$ score, one should find ideally as many models as possible plotted in the upper right corner of Figure \ref{fig:2d_gan_evaluation}. The discovered $PN$s generally perform with low $tp$ rates whereas the SGANs results are found closer towards the desired upper right area of the space. To counterfeit imbalances of $tp$ and $tp_{u}$, a score $s$ is introduced in Equation \ref{eq:score} which equals to the Euclidean distance between the origin $(0,0)$ and a model in the two-dimensional space represented by $tp$ and $tp_u$ weighted by the angle between the model vector in space and the reference one pointing to $(1,1)$. The intuition is that a model with low $tp$ and high $tp_{u}$ should not be considered better generalizing than a model with average, but balanced $tp$ and $tp_{u}$ scores. 
\begin{equation}\label{eq:score}
    \displaystyle 
    s(tp, tp_u) = \frac{tp + tp_u}{\sqrt{2}}
\end{equation}

\begin{figure}[ht]
\centering
    \includegraphics[width=230pt]{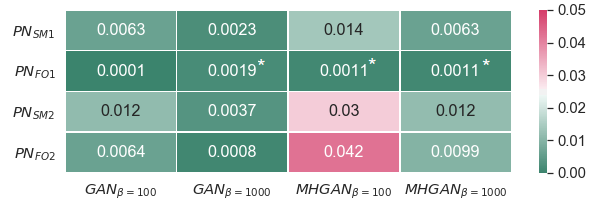}
  \caption{Confusion matrix of upper-tailed paired t-test p-values where the null hypothesis assumes that the mean of differences of $s$ of  SGAN sampling models are smaller or equal than the ones of the $PN$s. Values annotated with \mbox{*} indicate p-values resulting of a Wilcoxon signed-rank test.}
  \label{fig:t-test}
\end{figure}

One can statistically test if the mean differences of $s$ between an SGAN sampling model and a corresponding $PN$ are significantly higher. $16$ individual statistical tests are performed with each $15$ difference values comparing the playout outcomes of one $PN$ with one SGAN sampling architecture at a time. The Shapiro-Wilk test \cite{shapiro1965analysis} is an appropriate test for small sample sizes to assess if given values are normally distributed. An upper-tailed paired t-test \cite{mcdonald2009handbook} is performed when the Shapiro-Wilk test is passed, otherwise an upper-tailed Wilcoxon signed-rank test \cite{wilcoxon1992individual} assesses the mean difference. The resulting p-values are visualized in Figure \ref{fig:t-test}. Based on a standard confidence interval of $95\%$, the SGAN sampling models are in all cases statistically significant. To conclude, the AVATAR sampling methodology approximates the true number of system variants better, and detects a higher proportion of system variants and a higher ratio of unobserved system variants than the discovered process models (Q1-Q3).

\subsubsection{Generalization}\label{sec:Experimental-Evaluation:Generalization}
\begin{figure*}[!t]
    \includegraphics[width=\textwidth]{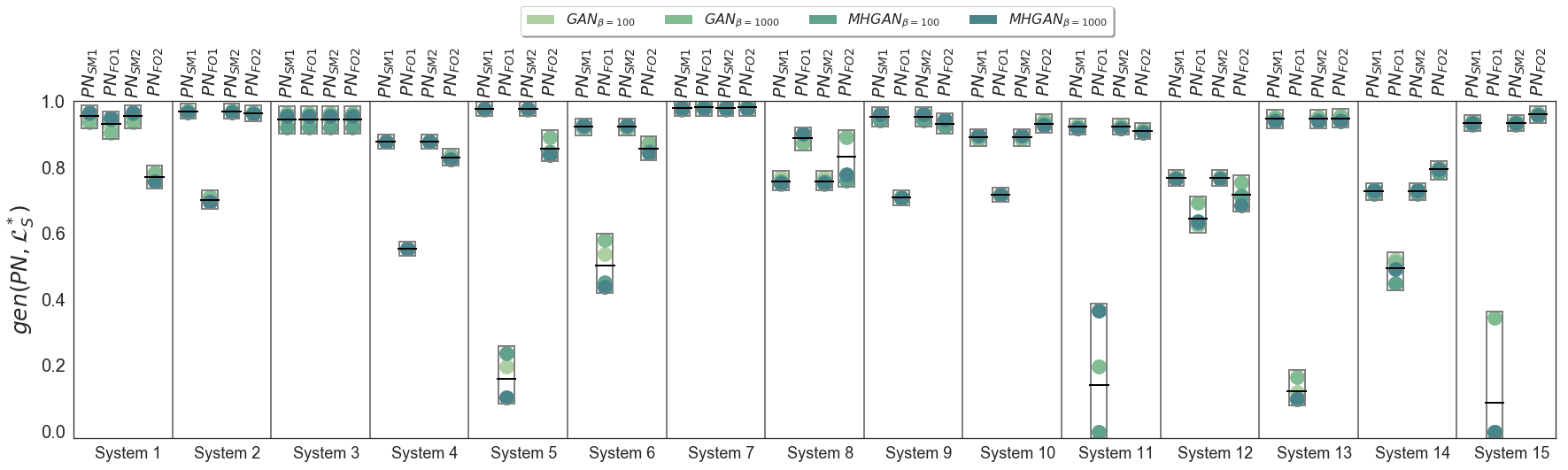}
  \caption{Obtained generalization scores following the proposed AVATAR methodology. The black horizontal lines indicate the arithmetic mean of the four generalization scores obtained for each $PN$.}
  \label{fig:generalization-results}
\end{figure*}  

For each $PN$, the corresponding four $\hat{\mathcal{V}}_S$ sets that are approximated by the SGAN models, are used to measure the generalization of the $PN$ as introduced in Equation \ref{eq:proposed-generalization} to answer Q4. The used fitness and precision functions are $fit^{\mathcal{L}}_t$ and $prec^{\mathcal{L}}_{etc}$ since those are the most established conformance measurement methods and some of the $PN$s are unsound requiring non-alignment-based metrics. However, any fitness and precision function can be used, as long as the definitions in Section \ref{sec:model-log-system-similarity} apply. For each $PN$, four generalization scores are reported corresponding to the four $\hat{\mathcal{V}}_S$ of the SGAN sampling models. The arithmetic mean of scores is supposed to represent the generalization of the $PN$. Figure \ref{fig:generalization-results} illustrates the obtained results which are closer to the expected and desired generalization scores than the ones in Section \ref{sec:pdandcc:results}.

Among all $PN$s of \textit{System 1}, $PN_{FO2}$ is the worst generalizing model. This is reflected by the generalization score visualized in Figure \ref{fig:generalization-results}. Similarly, $PN_{FO1}$ in \textit{System 2} is recognized as the least generalizing one which is correct since the model cannot play out any realistic variant. The AVATAR generalization scores for \textit{System 4} and \textit{7} are all identical which confirms the earlier observations. However, as expected, the scores are lower due to the observed $tp$ rate. This is caused by $fit^{\mathcal{L}}_t$ and $prec^{\mathcal{L}}_{etc}$, therefore more advanced $fit^\mathcal{L}$ and $prec^\mathcal{L}$ functions should be considered. For \textit{System 5}, both FO models are least generalizing, whereas $PN_{FO1}$ performs worse than $PN_{FO2}$. This is succesfully detected using the AVATAR methodology. 
The results for \textit{System 6} and \textit{2} are similar. $PN_{FO1}$ is succesfully detected as the least generalizing process model. Both SM $PN$s result in a high $tp$ rate and therefore in a higher generalization score. In comparison, $PN_{FO2}$ models more variants of $\mathcal{V}_u$, but at the cost of a low $tp$, thus a lower generalization score is obtained.

For high complex systems, all $PN_{FO1}$ models are detected to be least generalizing. The best score is obtained on \textit{System 15} and $PN_{FO2}$ which is a model with a comparatively high $tp$ and $tp_u$ rate, therefore confirming the proposed methodology. Notably, $PN$s discovered on \textit{System 14} generalize worse than $PN$s on other systems. 

To answer Q4, the proposed methodology accurately detects relative generalization when comparing process models of the same system and detects significant sets of unobserved variants. However,  the actual scores do not provide sufficient insight into the true generalization yet. 
This is caused by the drawbacks of $fit^{\mathcal{L}}_t$ and $prec^{\mathcal{L}}_{etc}$ \cite{Adriansyah2014}. With AVATAR, quantifying generalization is reduced to accurately measure $fit^{\mathcal{L}}$ and $prec^{\mathcal{L}}$. Based on the experiments, $fit^{\mathcal{L}}_t$ and $prec^{\mathcal{L}}_{etc}$ are anticipated to be replaced with more advanced methods. A future research study should address the evaluation using log fitness and log precision metrics beyond $fit^{\mathcal{L}}_t$ and $prec^{\mathcal{L}}_{etc}$ towards more precise and absolute generalization scores.

\section{Conclusion}\label{sec:conclusion}
This paper proposed a novel methodology to quantify the generalization of $PN$-based process models. Rather than measuring directly the extent to which a $PN$ represents the underlying unknown system, an SGAN is trained on an event log to obtain a set of the realistic variants of the system. The intuition relies on the objective of GANs to model an estimated distribution that converges towards the true underlying data distribution. The approach demonstrated statistical significance in unveiling unobserved realistic variants in a controlled experiment. Additionally, experiments have shown that generalization can be measured as the harmonic mean of log fitness and log precision on the estimated variant set of the system modeled by an SGAN.

In comparison to state-of-the-art methods that either try to measure the extent to which a process model generalizes the underlying system using the event log only or by interpreting structural properties of the $PN$, the proposed AVATAR methodology relies on a deep learning technique which has shown outstanding results in various computer vision and language modeling applications. Instead of relying solely on the observations, AVATAR approximates unobserved variants of a system by inferring knowledge beyond the event log. After revealing a set of unknown system variants, established log fitness and log precision metrics are leveraged to quantify generalization. This work demonstrates statistical significant results in obtaining unobserved variants using SGANs and succesfully detects  the relative generalization of process models. Thus, AVATAR is a step forward towards reliably measuring generalization. Besides a novel perspective on the metric of generalization, SGANs enable other real-world applications in the domain of process mining, such as for the analysis of black-box controller and software as in \cite{Theis2019c}, or to obtain robust and accurate simulation models.

In practice, access to the ground truth system is usually unavailable or restricted. However, to verify if newly yielded variants by AVATAR belong to the set of system variants, access to the ground truth system is required. This hinders the evaluation of the proposed approach when only real-world event logs are given. Future research needs to be conducted that focuses on the integration and verification of newly yielded variants by AVATAR into event log generation \cite{skydanienko2018tool} and process model enhancement methodologies.

Further future research is anticipated to be conducted in four directions. 
First, research studies are required that focus on more advanced log fitness and log precision methods towards precise and absolute generalization measurements.
Second, SGANs are in their infancy and efficient training with guaranteed satisfactory results is cumbersome. Hence, research is desired to focus on advancing SGAN architectures for the purpose of process mining.
Third, comprehensive elaborations on the impact of the size and bias of given event logs will mature the proposed methodology.
Finally, the approach demonstrated significant results on $15$ ground truth systems that were artificially created. To further underscore the significance of AVATAR, a comprehensive study on real-world systems is anticipated which might unveil further optimization possibilities.


%

\appendices
\section*{Appendix}
\subsection*{Appendix: Notations}
\begin{description}[style =standard, labelindent=0em , labelwidth=1.7cm, labelsep*=1em, leftmargin =!]
\item [$|\cdot|$] cardinality of a set
\item [$\lbrack:\rbrack$, $\lbrack;\rbrack$] column- and row-wise concatenation
\item [$a$] \textit{event}
\item [$\mathcal{A}$] finite set of all \textit{events}
\item [$\alpha$] acceptance probability in the MH algorithm
\item [$\beta$] inverse temperature RelGAN hyperparameter  
\item [$c$] trace
\item [$\mathcal{C}$] set of traces
\item [$c^i$] ith event instance of a trace $c$
\item [$d_K$] column dimension of the key $K$
\item [$D$] discriminator neural network
\item [$D_p$] discriminator trained on standard loss
\item [$D_r$] discriminator trained on relativistic loss
\item [$E$] event instance vector
\item [$e_1$] label of an event instance
\item [$\mathbf{E}$] expected value
\item [$\eta_1$] dependency threshold of FO
\item [$\eta_2$] length-one-loop threshold of FO
\item [$\eta_3$] length-two-loops threshold of FO
\item [$\varepsilon_1$] filtering hyperparameter of SM
\item [$\varepsilon_2$] parallelism hyperparameter of SM
\item [$f_{\theta_1}, f_{\theta_2}$] parametrized neural network functions
\item [$fp$] false positive rate
\item [$fit^\mathcal{L}$] log fitness
\item [$fit^\mathcal{L}_a$] token-based log fitness
\item [$fit^\mathcal{L}_t$] alignment-based log fitness
\item [$fit^S$] system fitness
\item [$\gamma(c)$] returns number of event instances of a trace $c$
\item [$G$] generator neural network
\item [$GAN_\beta$] naive SGAN-based sampling
\item [$gen$] generalization
\item [$gen_{ETM}$] ETM generalization function
\item [$gen(PN,\mathcal{L}_S^*)$] proposed AVATAR generalization
\item [$H$] set of heads
\item [$h$] head
\item [$k$] number of samples to draw per $GAN_\beta$
\item [$\kappa$] length of Markov chain
\item [$K$] key matrix
\item [$L$] loss function
\item [$L_D$] discriminator loss function
\item [$L_G$] generator loss function
\item [$\mathcal{L}$] event log
\item [$\mathcal{L}^*$] variant log
\item [$\mathcal{L}^+$] unique variant log
\item [$\mathcal{L}^+_e$] holdout unique variant log
\item [$\mathcal{L}^+_t$] training unique variant log
\item [$M$] memory matrix
\item [$MHGAN_\beta$] SGAN-based sampling with MH algorithm
\item [$\mu(\mathcal{L})$] maximum length of trace in an event log
\item [$\mathbb{N}$] proposal distribution of the MH algorithm
\item [$o_t$] output logit at time $t$
\item [$\omega(c)$] function mapping a trace $c$ to its variant
\item [$p$] probability density
\item [$\mathbb{P}$] probability distribution
\item [$PN$] Petri net process model
\item [$prec^\mathcal{L}$] log precision
\item [$prec^\mathcal{L}_a$] alignment-based log precision
\item [$prec^\mathcal{L}_{etc}$] log precision
\item [$prec^S$] system precision
\item [$\pi$] MH algorithm patience hyperparameter
\item [$\phi$] sigmoid function
\item [$\mathbb{Q}$] probability distribution
\item [$Q$] query matrix
\item [$R$] random variable of $S$ that can take on variants
\item [$S$] system
\item [$s(tp, tp_u)$] tradeoff function between $tp$ and $tp_u$
\item [$\sigma$] softmax function
\item [$t$] time
\item [$tp$] true positive rate
\item [$tp_S$] true positive rate over the $\mathcal{V}_S$
\item [$tp_o$] true positive rate over the $\mathcal{L}^+$
\item [$tp_u$] true positive rate over the $\mathcal{V}_u$
\item [$tp_e$] true positive rate over the $\mathcal{L}^+_e$
\item [$v$] variant
\item [$\mathcal{V}$] set of variants
\item [$\mathcal{V}_G$] set of variants that $G$ generates, estimator $\hat{\mathcal{V}}_S$
\item [$\mathcal{V}_{PN}$] set of all possible $PN$ variants, estimator $\hat{\mathcal{V}}_S$
\item [$\mathcal{V}_S$] set of all variants of a system $S$
\item [$\hat{\mathcal{V}}_S$] approximated set of all system variants
\item [$\mathcal{V}_u$] set of all unobserved variants of a system $S$
\item [$\hat{\mathcal{V}}_u$] est. set of all unobserved system variants
\item [$W$] weight matrix
\item [$x_r$] real sample
\item [$x_t$] sample at time $t$
\item [$x_g$] generated sample
\item [$Y$] values matrix
\item [$z$] noise sample
\item [$\mathbb{Z}$] noise distribution
\item [$x, n, y$] variables used in different contexts
\end{description}

\begin{table*}
\subsection*{Appendix: Process Discovery and Conformance Checking Results}
\resizebox{\textwidth}{!}{%
\begin{tabular}{|c|c|c|c|c|c|c|c|c|c|c|c|c|}
\rowcolor[HTML]{343434} 
{\color[HTML]{FFFFFF} \textbf{System}}              & {\color[HTML]{FFFFFF} \textbf{Process Discovery}} & {\color[HTML]{FFFFFF} \textbf{$fit^\mathcal{L}_t$}} & {\color[HTML]{FFFFFF} \textbf{$prec^\mathcal{L}_{etc}$}} & {\color[HTML]{FFFFFF} \textbf{$fit^\mathcal{L}_a$}} & {\color[HTML]{FFFFFF} \textbf{$prec^\mathcal{L}_a$}} & {\color[HTML]{FFFFFF} \textbf{$gen_{ETM}$}}   & {\color[HTML]{FFFFFF} \textbf{$|\hat{\mathcal{V}}_S|$}} & {\color[HTML]{FFFFFF} \textbf{$tp$}} & {\color[HTML]{FFFFFF} \textbf{$fp$}} & {\color[HTML]{FFFFFF} \textbf{$tp_{S}$}} & {\color[HTML]{FFFFFF} \textbf{$tp_{o}$}} & {\color[HTML]{FFFFFF} \textbf{$tp_{u}$}} \\
\rowcolor[HTML]{C0C0C0} 
\cellcolor[HTML]{C0C0C0}                             & $PN_{SM1}$ {[}0.1;0.3{]}                                 & 0.957                                 & 0.983                                  & 0.846                                 & 0.963                                  & \textbf{0.828}                        & 125                                        & 0.984                               & 0.016                               & 0.691                                     & 0.734                                       & 0.593                                      \\
\rowcolor[HTML]{C0C0C0} 
\cellcolor[HTML]{C0C0C0}                             & $PN_{FO1}$ {[}0.8;0.4;0.8{]}                             & 0.921                                 & 0.985                                  & 0.834                                 & 0.963                                  & \textbf{0.845}                        & 123                                        & 0.976                               & 0.024                               & 0.674                                     & 0.707                                       & 0.593                                      \\
\end{tabular}%
}
\caption{Evaluation of the discovered process models. The Process Discovery column indicates the applied process discovery algorithm along with the used hyperparameter combination where the first and second value correspond to $\varepsilon_1$ and $\varepsilon_2$, respectively, when using SM and where the first, second, and third value correspond to $\eta_1$, $\eta_2$, and $\eta_3$, respectively, when applying FO. N/A indicates that a particular score could not be obtained due to an unsound process model. Bold values indicate the conformance measures which were used for hyperparameter optimization of the corresponding discovery algorithm.}
\label{table:process-discovery}
\end{table*}

\begin{table*}
\subsection*{Appendix: AVATAR Sampling Results}
\centering
\resizebox{400pt}{!}{%
\begin{tabular}{|c|c|c|c|c|c|c|c|c|}
\rowcolor[HTML]{343434} 
{\color[HTML]{FFFFFF} \textbf{System}}              & {\color[HTML]{FFFFFF} \textbf{Sampling Approach}} & {\color[HTML]{FFFFFF} \textbf{$|\hat{\mathcal{V}}_S|$}} & {\color[HTML]{FFFFFF} \textbf{$tp$}} & {\color[HTML]{FFFFFF} \textbf{$fp$}} & {\color[HTML]{FFFFFF} \textbf{$tp_{S}$}} & {\color[HTML]{FFFFFF} \textbf{$tp_{o}$}} & {\color[HTML]{FFFFFF} \textbf{$tp_{u}$}} & {\color[HTML]{EFEFEF} \textbf{$tp_{e}$}} \\
\rowcolor[HTML]{C0C0C0} 
\cellcolor[HTML]{C0C0C0}                             & $GAN_{\beta=100}$                             & 176                                        & 0.6818                               & 0.3182                               & 0.6742                                     & 0.7823                                       & 0.4226                                      & 0.5385                                      \\
\rowcolor[HTML]{C0C0C0} 
\cellcolor[HTML]{C0C0C0}                             & $GAN_{\beta=1000}$                            & 188                                        & 0.4415                               & 0.5585                               & 0.4663                                     & 0.5645                                       & 0.2407                                      & 0.4615                                      \\
\rowcolor[HTML]{C0C0C0} 
\cellcolor[HTML]{C0C0C0}                             & $MHGAN_{\beta=100}$                              & 60                                         & 0.9833                               & 0.0167                               & 0.3315                                     & 0.4597                                       & 0.0370                                      & 0.0000                                      \\
\end{tabular}%
}
\caption{SGAN Sampling Results}
\label{table:gan_results}
\end{table*}

\begin{table*}
\subsection*{Appendix: AVATAR Generalization Scores}
\centering
\resizebox{420pt}{!}{%
\begin{tabular}{ccccccc}
\rowcolor[HTML]{333333} 
{\color[HTML]{FFFFFF} \textbf{System}}                                         & {\color[HTML]{FFFFFF} \textbf{PN}}                                 & \multicolumn{5}{c}{\cellcolor[HTML]{333333}{\color[HTML]{FFFFFF} \textbf{$gen(PN,\mathcal{L}_S^*)$}}}                                                                                                                                                                                                                                                                                                                                                                                       \\
\rowcolor[HTML]{525252} 
\multicolumn{1}{|c|}{\cellcolor[HTML]{525252}{\color[HTML]{FFFFFF} \textbf{}}} & \multicolumn{1}{c|}{\cellcolor[HTML]{525252}{\color[HTML]{FFFFFF} \textbf{}}} & \multicolumn{1}{c|}{\cellcolor[HTML]{525252}{\color[HTML]{FFFFFF} \textbf{$GAN_{\beta=100}$}}} & \multicolumn{1}{c|}{\cellcolor[HTML]{525252}{\color[HTML]{FFFFFF} \textbf{$GAN_{\beta=1000}$}}} & \multicolumn{1}{c|}{\cellcolor[HTML]{525252}{\color[HTML]{FFFFFF} \textbf{$MHGAN_{\beta=100}$}}} & \multicolumn{1}{c|}{\cellcolor[HTML]{525252}{\color[HTML]{FFFFFF} \textbf{$MHGAN_{\beta=1000}$}}} & \multicolumn{1}{c|}{\cellcolor[HTML]{525252}{\color[HTML]{FFFFFF} Mean}} \\
\rowcolor[HTML]{C0C0C0} 
\multicolumn{1}{|c|}{\cellcolor[HTML]{C0C0C0}}                                 & \multicolumn{1}{c|}{\cellcolor[HTML]{C0C0C0}$PN_{SM1}$}                       & \multicolumn{1}{c|}{\cellcolor[HTML]{C0C0C0}0.9537}                                            & \multicolumn{1}{c|}{\cellcolor[HTML]{C0C0C0}0.9371}                                             & \multicolumn{1}{c|}{\cellcolor[HTML]{C0C0C0}0.9636}                                              & \multicolumn{1}{c|}{\cellcolor[HTML]{C0C0C0}0.9671}                                               & \multicolumn{1}{c|}{\cellcolor[HTML]{C0C0C0}0.9554}                               \\
\end{tabular}%
}

\caption{AVATAR generalization results}
\label{table:generalization_results}
\end{table*}


\ifCLASSOPTIONcaptionsoff
  \newpage
\fi



%

\newpage
\bibliographystyle{IEEEtran}
\bibliography{IEEEabrv,references.bib}


\vskip -2\baselineskip plus -1fil

\begin{IEEEbiography}[{\includegraphics[width=1in,height=1.25in,clip,keepaspectratio]{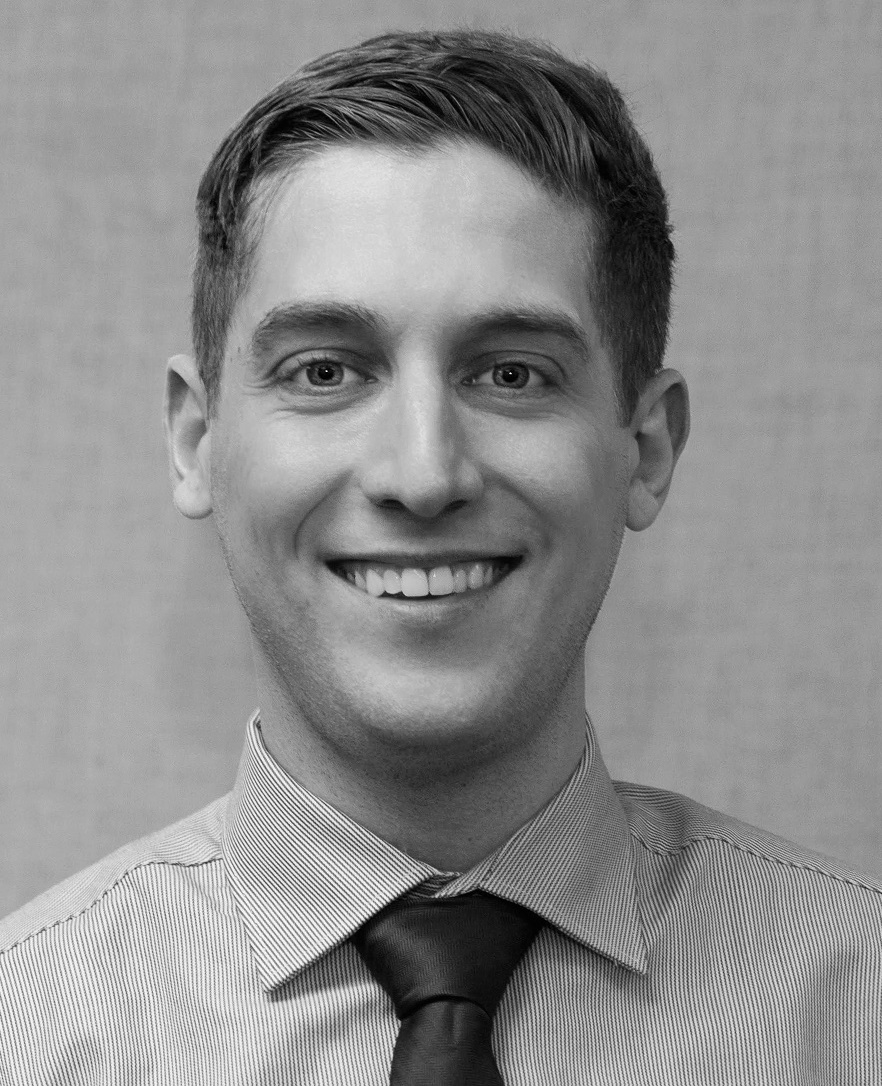}}]{Julian Theis} received the B.Eng. degree in Media Technology in 2014 and the M.Eng. degree with distinction in Media and Communications Technology in 2016, both from the RheinMain University of Applied Sciences, Wiesbaden, Germany. He is currently pursuing the Ph.D. degree with the Mechanical and Industrial Engineering Department, University of Illinois at Chicago. 

Before joining Prominent Laboratory, the university's foremost research facility in process mining, he was working as a software engineer for a leading company in media and workflow automation, broadcast management, and OSS software in Germany.
His research interests include process mining, its applications to industrial, IT, and business processes, and deep learning. 
\end{IEEEbiography}

\vspace{-1cm}

\begin{IEEEbiography}[{\includegraphics[width=1in,height=1.25in,clip,keepaspectratio]{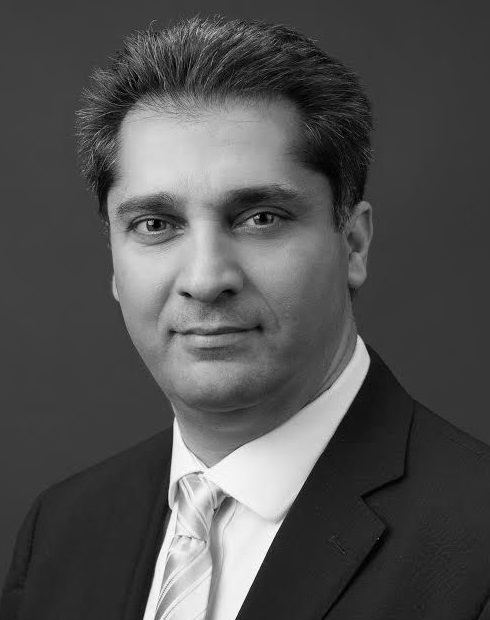}}]{Houshang Darabi} (S\textquotesingle98--A\textquotesingle00--M\textquotesingle10--SM\textquotesingle14) received the Ph.D. degree in industrial and systems engineering from Rutgers University, New Brunswick, NJ, USA, in 2000. 

He is currently a Professor with the Department of Mechanical and Industrial Engineering, University of Illinois at Chicago (UIC). He has been a contributing author of two books in the areas of scalable enterprise systems and reconfigurable discrete event systems. His research has been supported by several federal and private agencies, such as the National Science Foundation, the National Institute of Standard and Technology, the Department of Energy, and Motorola.  His current research interests include the application of data mining, process mining, and optimization in design and analysis of manufacturing, business, project management, and workflow management systems.
\end{IEEEbiography}


\vfill


\end{document}